\documentclass{article}
\PassOptionsToPackage{numbers, compress}{natbib}

\usepackage[preprint]{neurips_2024}

\usepackage[pagebackref,breaklinks,colorlinks,citecolor=cvprblue]{hyperref}
\usepackage{color, colortbl}
\usepackage{microtype}
\usepackage{graphicx}
\usepackage{subfigure}
\usepackage{booktabs} % for professional tables
\usepackage{wrapfig}
\usepackage{multirow}
\usepackage{graphicx}
\usepackage{pifont}% http://ctan.org/pkg/pifont for xmark and cmark
\usepackage{amsmath}
\usepackage{amssymb}
\usepackage{mathtools}
\usepackage{amsthm}
\usepackage{wrapfig}
\usepackage[textsize=tiny]{todonotes}
\usepackage{url}
\usepackage{caption}
\usepackage{makecell}
\usepackage[utf8]{inputenc} % allow utf-8 input
\usepackage[T1]{fontenc}    % use 8-bit T1 fonts
\usepackage{amsfonts}       % blackboard math symbols
\usepackage{nicefrac}       % compact symbols for 1/2, etc.
\usepackage{xcolor}         % colors
\theoremstyle{plain}

\theoremstyle{definition}

\theoremstyle{remark}

\newcommand{\myPara}[1]{\vspace{.05in}\noindent\textbf{#1}}

\newcommand{\eg}{\emph{e.g.}}

\newcommand{\ie}{\emph{i.e.}}

\definecolor{darkblue}{rgb}{0, 0.12, 0.55}
\definecolor{darkgreen}{rgb}{0, 0.55, 0.12}
\definecolor{darkred}{rgb}{0.6,0,0}
\definecolor{darkgreen}{rgb}{0,0.6,0}
\definecolor{Gray}{gray}{0.9}
\definecolor{lightblue}{rgb}{0.68, 0.85, 0.9}
\definecolor{cvprblue}{rgb}{0.21,0.49,0.74}

\title{Open-Vocabulary Segmentation with \\Unpaired Mask-Text Supervision}

\author{
Zhaoqing Wang$^1$ \,\, Xiaobo Xia$^1$ \,\, Ziye Chen$^2$ \,\, Xiao He$^3$ \,\, Yandong Guo$^3$ \\
\textbf{Mingming Gong}$^2$ \,\, \textbf{Tongliang Liu}$^1$ \\
$^1$ Sydney AI Centre, The University of Sydney \\ $^2$ Melbourne Centre for Data Science, The University of Melbourne \,\,\\
$^3$ AI2Robotics \\
\texttt{zwan6779@uni.sydney.edu.au; mingming.gong@unimelb.edu.au} \\
\texttt{tongliang.liu@sydney.edu.au} \\
\url{https://github.com/DerrickWang005/Unpair-Seg.pytorch}
}

\begin{document}
\maketitle

\begin{abstract}
\label{sec:abstract}
    Current state-of-the-art open-vocabulary segmentation methods typically rely on image-mask-text triplet annotations for supervision. However, acquiring such detailed annotations is labour-intensive and poses scalability challenges in complex real-world scenarios. While existing weakly-supervised approaches leverage image-text pairs to reduce the expansive annotation cost, the lack of mask supervision makes it difficult for the model to locate multiple instances and accurately group pixels with similar semantics, significantly hampering versatility and performance. In this paper, we introduce Unpair-Seg, a novel weakly-supervised open-vocabulary segmentation framework that learns from unpaired image-mask and image-text pairs, which can be independently and efficiently collected. Unpair-Seg initially predicts a set of binary masks and generates pseudo labels by identifying confident pairs of masks and text entities. We then train a feature adapter to align region embeddings with text embeddings based on these pseudo labels, achieving open-vocabulary segmentation. However, the inherent noise in the mask-entity correspondence poses a challenge to obtaining reliable pairs. To address this, we employ a vision-language large model to re-caption the input images and extract precise entities, and we design a multi-scale matching strategy to reduce noisy mask-entity pairs. Our Unpair-Seg framework demonstrates impressive performance, achieving 14.6\% and 19.5\% mIoU on the ADE-847 and PASCAL Context-459 datasets, significantly narrowing the gap between fully-supervised and weakly-supervised methods.
    % Our source code is available at \url{https://github.com/DerrickWang005/Unpair-Seg.pytorch}.

\end{abstract}

\section{Introduction}\label{sec:intro}
Open-vocabulary segmentation refers to the segmentation and categorisation of objects from an expansive and unrestricted vocabulary, even though the object categories within the vocabulary are not encountered during training~\cite{li2022languagedriven,ding2022decoupling}. Compared to traditional closed-vocabulary segmentation~\cite{zhang2021k,cheng2022masked}, which depends on predefined training categories and cannot recognise absent categories, open-vocabulary segmentation can segment any category with arbitrary text descriptions. This innovative segmentation paradigm has garnered considerable attention~\cite{xu2022simple,liang2023open} and opened up numerous potential applications~\cite{zou2023generalized,zou2023segment,you2023ferret}.

Cutting-edge approaches in open-vocabulary segmentation typically leverage the supervision of triplet annotations composed of images, masks, and corresponding texts~\cite{yu2023convolutions}. It is worth noting that strict alignment between each mask and text results in an expensive annotation cost. To mitigate this, existing weakly-supervised methods are proposed to use image-text pairs, \ie, text supervision~\cite{xu2022groupvit,cha2023learning,xu2023learning}. However, learning with this type of supervision, the model is suboptimal to dense prediction because it is difficult to understand complex spatial relations. Furthermore, this type of supervision lacks positional information, making it challenging for the model to distinguish multiple instances with the same semantic class. The issues mentioned above severely limit these weakly-supervised methods' versatility and segmentation performance.

\begin{figure}[t]
    \centering
    \includegraphics[width=1.0\textwidth]{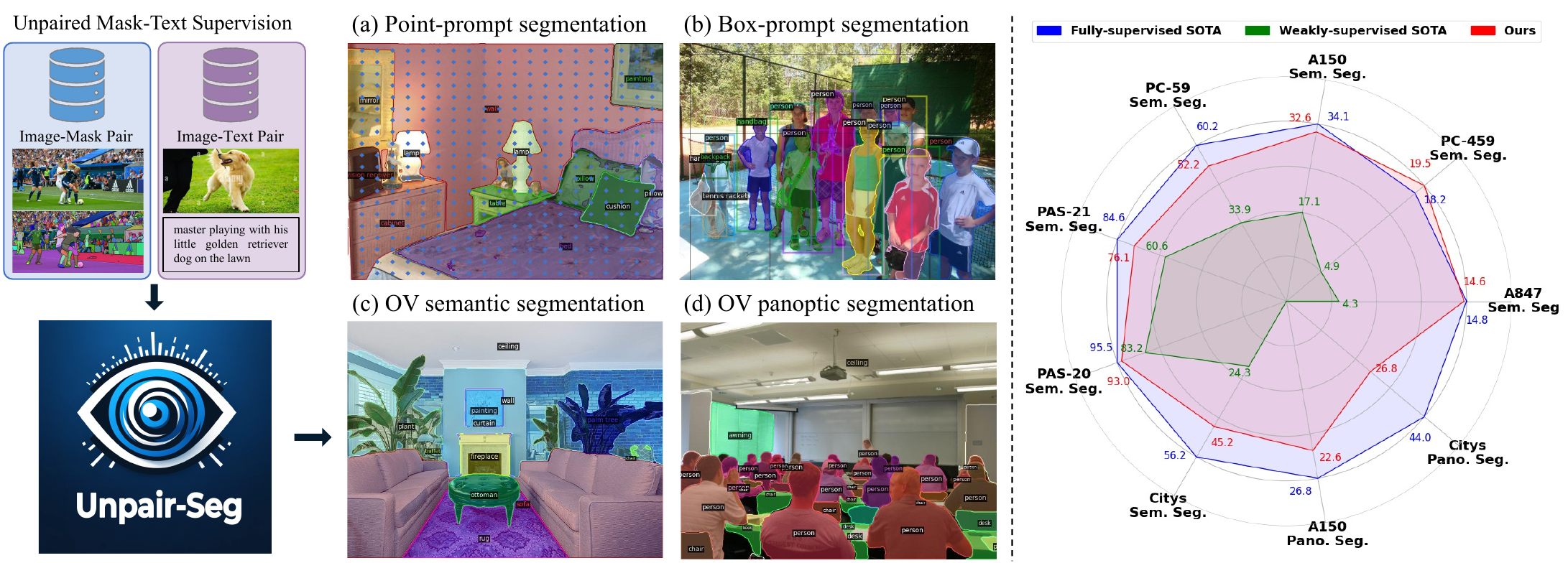}
    \captionof{figure}{Unpair-Seg framework directly learns from \textbf{unpaired mask-text supervision}. Unlike labour-intensive image-mask-text annotations, independent image-mask and image-text pairs are more accessible to collect. With a single set of weights, Unpair-Seg excels at various image segmentation tasks, including point-prompt, box-prompt, open-vocabulary semantic and panoptic segmentation. Extensive experimental results demonstrate that our method significantly narrows the gap between fully-supervised and weakly-supervised approaches.
    }
    \vspace{-0.4cm}
    \label{fig:teaser}
\end{figure}

% In this paper, to reduce the annotation expense and develop scalable open-vocabulary segmentation, we propose a weakly-supervised open-vocabulary segmentation framework named Uni-OVSeg.
% In essence, our Uni-OVSeg eliminates the requirement for triplet annotations and allows independent image-mask and image-text pairs, where masks and text can be unpaired.
% In essence, our Uni-OVSeg eliminates the requirement for triplet annotations and allows independent image-mask and image-text pairs.
% Evidently, the supervision in Uni-OVSeg is weaker than the previous triplet supervision and easier to collect, since it liberates correspondence between masks and text.
% In further detail, Uni-OVSeg works in an end-to-end manner and comprises two primary branches: mask segmentation and mask-text alignment. 
% Mask segmentation showcases the model's spatial understanding by grouping semantically similar pixels. Mask-text alignment demonstrates the model's ability to align pixel-wise feature representation with the structure of text, endowing pixel-wise features with the extensibility of text. This integration helps to achieve open-vocabulary segmentation.
Motivated by this, we propose an advanced weakly-supervised open-vocabulary segmentation framework, Unpair-Seg, to reduce the annotation expense while significantly enhancing performance. In essence, we liberate the strict correspondence between masks and texts by using independent image-mask and image-text pairs. These two types of pairs can be easily collected from different sources, as illustrated in Figure~\ref{fig:teaser}, such as SA-1B~\cite{kirillov2023segment} for image-mask pairs and LAION-400M~\cite{schuhmann2021laion} for image-text pairs. By resorting to independent image-mask and image-text pairs, our Unpair-Seg has a strong segmentation ability to group semantically similar pixels and align regional embeddings with entity embeddings of texts into the same representation space, achieving open-vocabulary segmentation.

Technically, when presented with an image-mask pair, we train a visual prompt encoder, a pixel decoder, and a mask decoder to generate a set of binary masks based on different types of visual prompts (\eg, points and bounding boxes). The collected image-text pair often contains certain texts that do not match the image~\cite{thomee2016yfcc100m}, leading to an incorrect correspondence between masks and entities. We employ the vision-language large model (\ie, InternLM-XComposer~\cite{internlmxcomposer2}) to re-caption each input image and extract more precise entities from the text description. To train mask-entity alignment, we introduce bipartite matching to exploit confident pairs of predicted masks and entities in the CLIP embedding space~\cite{radford2021learning}, generating pseudo labels. Due to the inherent noise in the correspondence between vision and language, we improve the quality of visual embedding by ensembling multi-scale clip features, thereby stabilising the matching process. Afterwards, given these pseudo labels, we train a feature adapter to align regional embeddings of predicted masks and entity embeddings of text descriptions. During the inference phase, a zero-shot classifier, built by embedding target dataset category names, assigns a category to each predicted mask, enabling the system to segment objects across an open vocabulary.

Before delving into details, we summarise our contributions in the following aspects:
\begin{itemize}
    \item We introduce Unpair-Seg, a novel weakly-supervised open-vocabulary segmentation framework that significantly reduces the need for costly triplet annotations, making open-vocabulary segmentation more accessible and scalable.
    \item Considering the inherent noise in the vision-language correspondence, we employ the vision-language large model to extract precise entities from input images and design a multi-scale matching strategy to exploit confident pairs of masks and entities.
    \item With one suit of weights, we achieve impressive results of 22.6\% PQ and 19.5\% mIoU on the challenging ADE20k panoptic segmentation and PASCAL Context-459 semantic segmentation, respectively. Comprehensive ablation studies and discussions are also provided.
\end{itemize}
\section{Related works}
\label{sec:related}

\myPara{Generic segmentation.}
Given an image, segmentation of specific visual concepts has remained an ongoing research topic in computer vision, as indicated by the extensive literature on it~\cite{long2015fully,he2017mask,kirillov2019panoptic}. Generic segmentation mainly includes semantic segmentation~\cite{long2015fully,zhang2018context,yuan2021ocnet,cheng2021per}, instance segmentation~\cite{he2017mask,bolya2019yolact,chen2019hybrid}, and panoptic segmentation~\cite{kirillov2019panoptic,cheng2020panoptic,wang2020axial,cheng2020panoptic}, related to different levels of granularity. In more detail, semantic segmentation~\cite{fu2019dual,cheng2021per,zhang2022segvit,liang2022gmmseg} aims to assign a label to each pixel of the input image according to their respective semantic classes. In addition, instance segmentation~\cite{tian2020conditional,wang2021solo,tian2022instance} attempts to distinguish different object instances of the same semantic class. Panoptic segmentation~\cite{wang2021max,yu2022k,yu2022cmt,chen2023generalist} combines the characteristics of semantic segmentation and instance segmentation. Following a close-vocabulary assumption, previous works can only predict predefined object categories. In this paper, we aim to build an advanced open-vocabulary segmentation system which can categorise objects and stuff from an open set of vocabulary in the real world.

\myPara{Vision foundation models.}
Recent advancements in visual foundation models have diversified optimisation techniques across various learning paradigms. These developments range from vision-only pretraining~\cite{he2022masked,he2020momentum} to joint vision-language pre-training~\cite{radford2021learning,jia2021scaling,yu2022coca}, and extend to multi-modal frameworks that integrate visual prompting~\cite{alayrac2022flamingo}. A prime example of this evolution is SAM~\cite{kirillov2023segment}, which shows the potential of extensive training for general segmentation, offering impressive generalisability and scalability. Despite its impressive capabilities, SAM cannot categorise predicted masks into different semantic classes, which is limited by the supervision of the image-mask pairs. More recently, Semantic-SAM~\cite{li2023semantic} unifies different sources of human-annotated segmentation datasets and augments SAM by adding semantic labels and increased levels of granularity. Our work aims to develop a more flexible vision foundation model, which can be trained with unpaired mask-text supervision (\eg, independent image-mask and image-text pairs) and can be easily adapted to different segmentation tasks.

\myPara{Open-vocabulary segmentation.}
Open-vocabulary segmentation counters the constraints of closed-vocabulary segmentation by allowing the segmentation of a diverse range of classes, even those unseen during training~\cite{ghiasi2022scaling,xu2022simple,zhang2023simple,xu2023side}. Existing works~\cite{ding2023open,yu2023convolutions,xu2023open} leverage the pretrained vision-language models~(\eg, CLIP~\cite{radford2021learning} and ALIGN~\cite{jia2021scaling}) to perform open-vocabulary segmentation. Most open-vocabulary segmentation methods commonly utilise human-annotated supervision (\ie, the image-mask-text triplets) to generalise the capability of vision-language models from the image level to the pixel level. To reduce the dependency on this labour-intensive supervision, some weakly-supervised methods are proposed to use only text supervisions~\cite{luo2023segclip,xu2023learning}. They learn to group image regions into shaped segments, but struggle to distinguish different instances with the same semantic class and the segmentation performance is unsatisfactory~\cite{zhou2022extract,xu2022groupvit}. This dilemma drives our pursuit of a more advanced open-vocabulary segmentation framework, where the aim is to enjoy as low an annotation cost as possible and simultaneously achieve significant performance.

\begin{figure*}[t]
    \begin{center}
    \centerline{
        \includegraphics[width=\textwidth]{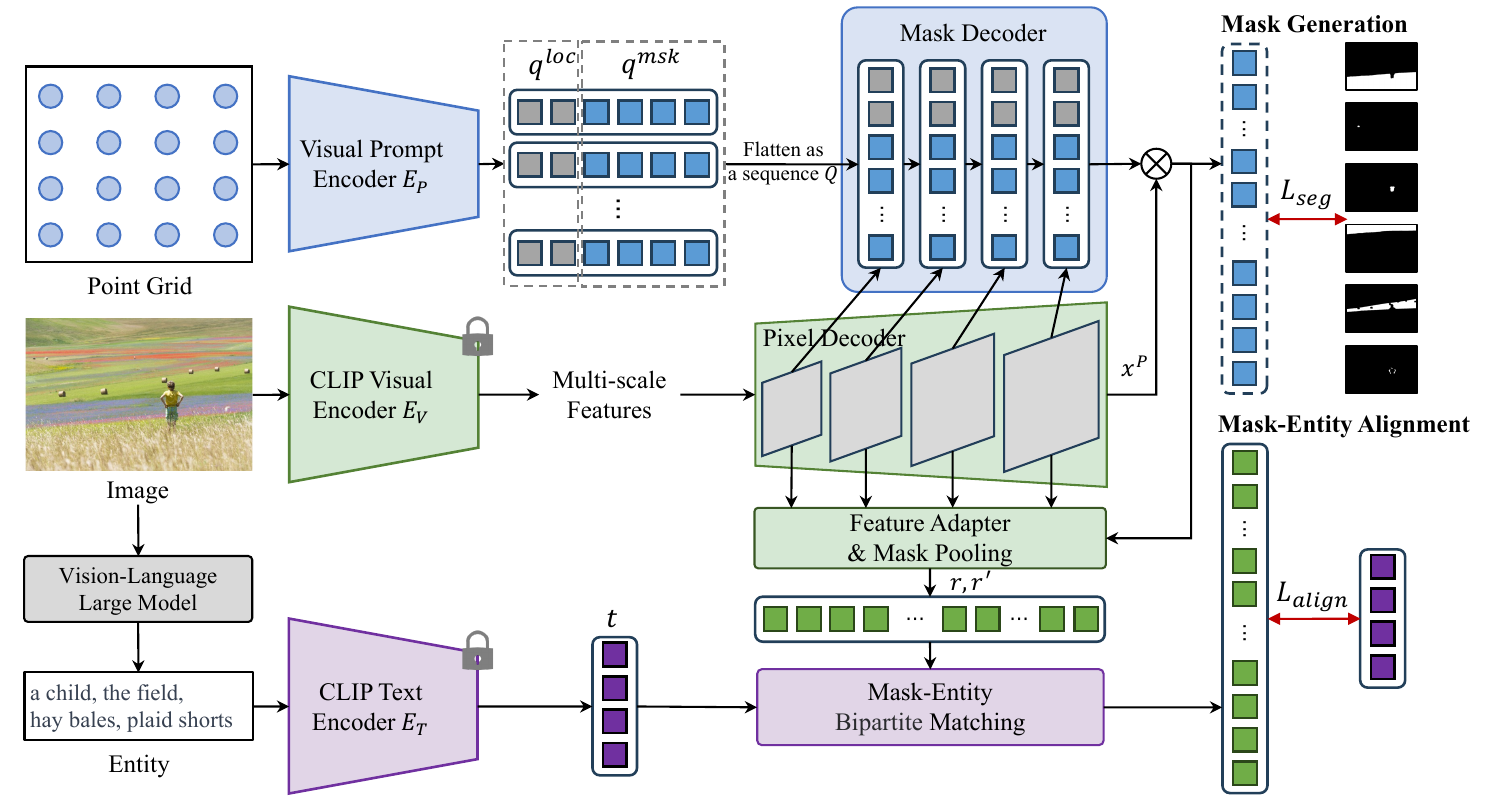}
    }
    \vspace{-0.2cm}
    \caption{\textbf{Overview of the proposed Unpair-Seg framework.}
    Our framework consists of two stages, including mask generation and mask-entity alignment. Given image-mask pairs, we first train a prompt encoder, pixel decoder, and mask decoder for binary mask generation. Subsequently, when presented with image-text pairs, a feature adapter is optimized to align regional embeddings of predicted masks and entity embeddings of text descriptions. A mask-entity bipartite matching is designed to assign the corresponding mask prediction for each entity. CLIP visual and text encoders are frozen. Visual prompts using boxes are omitted for simplicity.
    }
    \label{fig:architecture}
    \end{center}
    \vspace{-0.8cm}
\end{figure*}

\section{Method}
\label{sec:method}
In this section, we first define the problem of open-vocabulary segmentation in Section~\ref{subsec:prob_def}. We then introduce a straightforward baseline in Section~\ref{subsec:base}. Finally, we present our proposed Unpair-Seg framework in Section~\ref{subsec:uniovseg}, including an overview, mask generation, mask-entity alignment, and open-vocabulary inference.

\subsection{Problem definition}
\label{subsec:prob_def}
Given an image $\mathbf{I}\in\mathbb{R}^{H\times W\times 3}$, where $H$ and $W$ denote the height and width of the image respectively, open-vocabulary segmentation aims to segment the image into a set of masks with associated semantic classes:
\begin{equation}
    \{\mathbf{y}_i\}_{i=1}^k=\{(\mathbf{m}_i,\mathbf{c}_i)\}_{i=1}^k.
\end{equation}
The $k$ masks $\mathbf{m}_i\in\{0,1\}^{H\times W}$ include the associated ground truth class $\mathbf{c}_i$~\cite{yu2023convolutions}.
Unlike traditional image segmentation tasks~\cite{long2015fully,he2017mask,kirillov2019panoptic}, open-vocabulary segmentation is more challenging because inference classes are not observed during training.
During the evaluation, the test categories $\mathbf{C}_\text{test}$ are different from $\mathbf{C}_\text{train}$, which contain novel categories not seen in training, \ie, $\mathbf{C}_\text{train}\neq\mathbf{C}_\text{test}$.
Different from previous works~\cite{xu2022groupvit,xu2023open}, in our setting, no paired human-annotated mask and semantic category is provided in advance for any training image.
We assume that the category names of $\mathbf{C}_\text{test}$ are available, represented in the form of natural language.

\subsection{Baseline}
\label{subsec:base}
We introduce a straightforward baseline using the knowledge of image-text and image-mask pairs. Specifically, we employ a CLIP model as the visual and text encoder, which is trained on a large amount of image-text pairs. Afterwards, we use the image-mask pairs to obtain a branch of mask generation, predicting a set of binary masks. To perform open-vocabulary segmentation, we crop and pool the CLIP image features based on these predicted masks, which are further classified by the CLIP text embeddings. Although this straightforward baseline enables open-vocabulary segmentation, it exhibits a noticeable knowledge gap between the image-level and pixel-level tasks.

\subsection{Unpair-Seg framework}
\label{subsec:uniovseg}
\myPara{Overview.}
Our novel framework, Unpair-Seg, for weakly-supervised open-vocabulary segmentation, is depicted in Figure~\ref {fig:architecture}.
On a macro level, our framework contains a CLIP model to extract features of both images and text descriptions.
Using the image-mask pairs, our framework employs a mask generation branch. This branch includes a visual prompt encoder, a pixel decoder, and a mask decoder, all of which work together to predict a set of binary masks for an input image.
With the image-text pairs, mask-entity bipartite matching exploits confident pairs between predicted masks and entities in text descriptions.
Afterwards, we adopt a multi-scale feature adapter to enhance regional embeddings of mask predictions, which are further aligned with associated entity embeddings based on the confident pairs.
Finally, we perform open-vocabulary segmentation with the parts mentioned above. More details can be found in Appendix~\ref{sec:detail}.

\myPara{Mask generation.}
Given an input image $\mathbf{I}$, mask generation aims to predict a set of binary masks $\mathbf{m}$.
These masks represent groups of semantically similar pixels at various image locations.
To effectively associate these masks with visual prompts (\eg, points and boxes), we adopt an architecture inspired by~\cite{cheng2022masked,kirillov2023segment}, employing a visual prompt encoder, a pixel decoder and a query-based mask decoder.
Point $(x, y)$ and box $(x, y, w, h)$ prompts are transformed into unified anchor boxes.
Each point is converted into an anchor box with a predefined width $w^{'}$ and height $h^{'}$, effectively approximating the location of points.
To address the challenge of predicting masks with variable granularities, as depicted in Fig.~\ref{fig:architecture}, we encode each point into two positional embeddings $\mathbf{q}^{loc}_{n}\in\mathbb{R}^{2 \times dim}$ using sinusoidal encoding~\cite{vaswani2017attention}, and combine them with $M$ distinct mask embeddings $\mathbf{q}^{msk}\in\mathbb{R}^{M \times dim}$.
Each mask embedding is a learnable vector corresponding to a specific granularity level.
For box prompts, the process involves encoding into two position embeddings followed by concatenation with a singular mask embedding, as each box is linked to a unique object.
Afterward, we represent $N$ input location prompts as a set of query embeddings $\mathbf{Q}={ \mathbf{q}_1, ..., \mathbf{q}_{n} }$. For the $i$-th query,
\begin{equation}
    \mathbf{q}_i = Concat(\mathbf{q}^{loc}_{i}; \mathbf{q}^{msk}) + \mathbf{q}^{type} + \mathbf{q}^{feat},
\end{equation}
where $\mathbf{q}^{type}\in\mathbb{R}^{dim}$ denotes the query type, chosen from either the point or the box.
That $\mathbf{q}^{feat}\in\mathbb{R}^{dim}$ is the content embedding sampled from visual features.
For box prompts, we sample visual features based on the centre of each box.

Subsequently, to capture the multi-scale information, a pixel decoder equipped with multi-scale deformable attention~\cite{zhu2020deformable} is used and outputs enhanced pixel features $\mathbf{x}^{p}$.
These features are fed into the query-based mask decoder in conjunction with the query embeddings to predict binary masks.
Each layer of the mask decoder comprises a masked cross-attention layer~\cite{cheng2022masked}, a self-attention layer~\cite{vaswani2017attention}, and a feed-forward network.
The binary masks for each visual prompt $\mathbf{m}_n$ are derived through a matrix multiplication between the query embeddings and pixel features,
\begin{equation}
    \mathbf{m}_n = Sigmoid(\mathbf{q}^{msk}_n \cdot \mathbf{x}^{p}),
\end{equation}
where $\mathbf{m}_n\in\mathbb{R}^{M\times H\times W}$ and $Sigmoid(\cdot)$ is the sigmoid function to normalize the mask values into $[0, 1]$.
% Considering that the input points may belong to multiple masks simultaneously,
An input point is likely to exist in different granularity masks simultaneously.
Given one point prompt, SAM~\cite{kirillov2023segment} selects the prediction with the minimum cost to the ground truth for loss calculation, which cannot effectively capture the inherent ambiguity of point prompts.
To model this property and accurately predict multi-granularity masks, in contrast, we design a many-to-many matching strategy to automatically associate the ground-truth masks with the mask predictions of each point, allowing a more adequate use of image-mask pairs.
Once matching, we compute the segmentation loss for each mask prediction $\mathbf{m}_i$ with the corresponding ground-truth mask $\mathbf{y}^{msk}_j$. Besides, we also estimate the quality of mask generation by regressing the intersection-over-union $\mathbf{y}^{iou}_{i,j}$ between the mask prediction $\mathbf{m}_i$ and ground-truth mask $\mathbf{y}^{msk}_j$,
\begin{equation}
    % \mathcal{L}_{seg}(\mathbf{m}_i, \mathbf{y}^{msk}_j) = \lambda_{bce}\cdot\mathcal{L}_{bce}(\mathbf{m}_i, \mathbf{y}^{msk}_j) + \lambda_{dice}\cdot\mathcal{L}_{dice}(\mathbf{m}_i, \mathbf{y}^{msk}_j) + \lambda_{iou}\cdot\mathcal{L}_{iou}(\mathbf{q}_i, \mathbf{y}^{iou}_{i,j}),
    \mathcal{L}_{seg} = \lambda_{bce}\cdot\mathcal{L}_{bce}(\mathbf{m}_i, \mathbf{y}^{msk}_j) + \lambda_{dice}\cdot\mathcal{L}_{dice}(\mathbf{m}_i, \mathbf{y}^{msk}_j) + \lambda_{iou}\cdot\mathcal{L}_{iou}(\mathbf{q}_i, \mathbf{y}^{iou}_{i,j}),
\end{equation}
where $\lambda_{bce}$, $\lambda_{dice}$, and $\lambda_{iou}$ are three hyper-parameters to balance the binary cross-entropy loss $\mathcal{L}_{bce}$, the dice loss $\mathcal{L}_{dice}$ and the quality loss $\mathcal{L}_{iou}$.
% Note that we apply no penalty to the remaining unmatched mask predictions, which ensures that multiple granularity masks can be predicted for each point.
\begin{wrapfigure}[17]{r}{0.5\textwidth}
  \centering
  % \vspace{-0.2cm}
  \caption{Comparison between raw and improved text descriptions.
  ``\textit{Misalign.}'', ``\textit{Deficient.}'', and ``\textit{Missing.}'' denote text-image misalignment, deficient description, and missing text description.
  % We can accurately extract entities from improved descriptions.
  \includegraphics[width=0.48\textwidth]{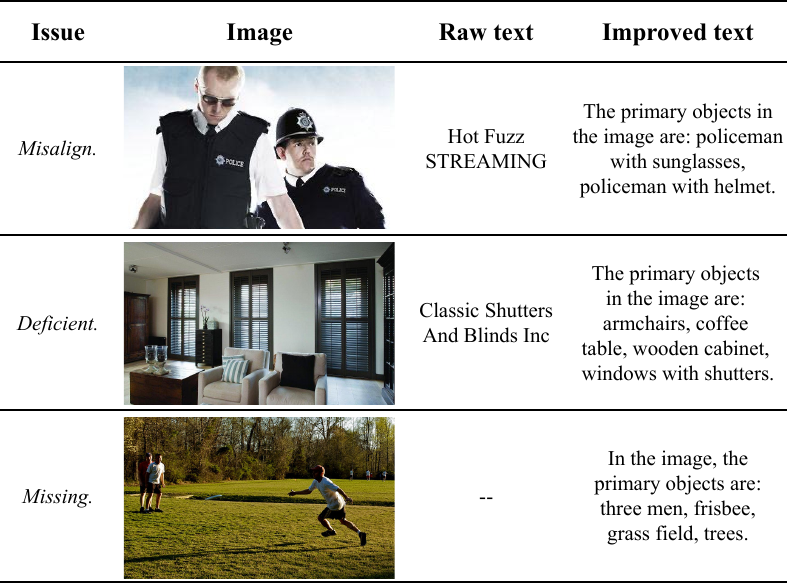}
  }
  \label{tab:cap}
\end{wrapfigure}
Note that we do not apply any penalty to the remaining unmatched predicted masks. In doing so, we ensure the model to predict multiple granularity masks of each input point prompt.

\myPara{Mask-entity alignment.}
In order to enable the model to categorize the predicted masks from a wide range of vocabulary, we need to establish a connection between objects in the image and entities in the text description based on image-text pairs. However, these pairs often present challenges such as text-image misalignment and incomplete descriptions. Some images even lack text descriptions, making it especially difficult to align the predicted masks with the corresponding entities directly.

To address these challenges, we utilize the vision-language large model (VLLM), InternLM-XComposer~\cite{internlmxcomposer2}, to re-caption each input image. Using a prompt like ``\textit{You are a proficient annotator. Please output primary objects in the input image without duplication. These output objects should be connected by commas.}'', we can accurately identify entities from the text descriptions, as illustrated in Figure~\ref{tab:cap}.

Following this, each identified entity and the prompts used in previous work~\cite{gu2021open} are input into the CLIP text encoder to obtain the text embedding $\mathbf{t}_{k}$. Once a set of binary masks $\mathbf{m}$ is generated from the input image, we obtain the region embeddings $\mathbf{r}_i$ by using a mask pooling layer $P$ and the CLIP visual projector $F_{v}$. That is,
\begin{equation}
    \mathbf{r}_i = F_{v}(P(\mathbf{x}^{v}, \textbf{m}_i)),
\end{equation}
where $\mathbf{x}^{v}$ denotes the visual features extracted from the CLIP visual encoder.
The scale of objects within images often varies significantly. To stabilize the process of matching mask entities across multi-resolution images, we extract regional embeddings from multi-scale dense visual features and then ensemble these regional embeddings to enhance the quality,
\begin{equation}
    \mathbf{\hat{r}}_{i} = \frac{1}{S}\sum_{s=1}^{S}\mathbf{r}_{i,s},
\end{equation}
where $S$ denotes the number of scales. Afterwards, we calculate a cost matrix $\mathbf{\delta}$ between the region embeddings $\mathbf{\hat{r}}$ and text embeddings $\mathbf{t}$,
\begin{equation}
    \mathbf{\delta}_{i,j}=\frac{\exp(\mathbf{\delta}^{'}_{i,j})}{\sum_{j}\exp(\delta^{'}_{i,j})},\quad \mathbf{\delta}^{'}_{i,k} = 1 - \frac{\mathbf{r}_i \cdot \mathbf{t}_k}{\|\mathbf{r}_i\|_2 \|\mathbf{t}_k\|_2},
\end{equation}
where $\mathbf{\delta}^{'}_{i,k}$ denotes the reverse cosine similarity between the $i$-th region embedding and the $k$-th text embedding.
With this matrix, we apply the bipartite matching algorithm~\cite{karp1990optimal} to obtain the confident pairs.
Subsequently, we input the multi-scale visual features obtained from the pixel decoder and the CLIP visual features into a feature adapter to obtain the advanced region embeddings $\mathbf{r}^{'}_i$. More details about the feature adapter can be found in Appendix~\ref{sec:detail}. We then compute the cosine similarity loss for each paired region and text embedding:
\begin{equation}
    \mathcal{L}_{align}(\mathbf{r}^{'}_i, \mathbf{t}_k) = 1 - \frac{\mathbf{r}^{'}_i \cdot \mathbf{t}_k}{\|\mathbf{r}^{'}_i\|_2 \|\mathbf{t}_k\|_2}.
\end{equation}

\myPara{Open-vocabulary inference.}
During inference, we start with the test categories $\mathbf{C}_\text{test}$. Then, we engage in prompt engineering~\cite{ghiasi2022scaling} using the CLIP text encoder to extract text embeddings for open-vocabulary segmentation. For each input image, we use a uniform point grid as the visual prompt to predict a set of binary masks. We next calculate the cosine similarity with the text embeddings to predict the category with the highest similarity to the label of the corresponding mask.

\section{Experiments}
\label{sec:experiment}

\subsection{Implementation details}
\myPara{Datasets.}
During training, we randomly sample the 50\% subset from the SA-1B dataset~\cite{kirillov2023segment}, which contains $\sim 5$ million images and $\sim 0.5$ billion masks.
Although this supervision provides diverse binary masks, it lacks the semantic class for each mask.
Besides, we collect about 130k images~\cite{caesar2018coco,cordts2016cityscapes} and adopt a vision-language large model~\cite{internlmxcomposer2} to generate captions, forming image-text pairs. Afterwards, we extract entities from these text descriptions.

\begin{table}[!t]
    \centering
    \caption{
        \textbf{Open-vocabulary semantic segmentation performance.}
        We mainly compare with the fully-supervised and weakly-supervised methods.
        ``COCO S.'', ``COCO P.'', and ``COCO C.'' denote the COCO stuff, panoptic, and caption datasets.
        ``O365'' denotes the Object 365 dataset. ``M. 41M'' denotes the merged 41M image dataset.
        We report mIoU for all datasets.
    }
    % \vspace{-0.3cm}
    \label{tab:ov_semseg}
    \setlength{\tabcolsep}{1.2mm}{
    % \small
    \footnotesize
    \begin{tabular}{l|c|ccccccc}
    \toprule
    \hline
    \multirow{2}{*}{\textbf{Method}} & \multirow{2}{*}{\textbf{Training Data}} & \textbf{A-847} & \textbf{PC-459} & \textbf{A-150} & \textbf{PC-59} & \textbf{PAS-21} & \textbf{PAS-20} & \textbf{Citys} \\
    \cline{3-9}
     &  & \multicolumn{7}{c}{\textbf{mIoU (\%)}} \\
    \hline
    \multicolumn{9}{l}{\textit{\textbf{Fully-supervised method (image-text-mask pair)}}} \\
    \hline
    % SPNet & VOC & - & - & - & 24.3 & 18.3 & - & - \\
    % ZS3Net & VOC & - & - & - & 19.4 & 38.3 & - & - \\
    % LSeg & VOC & - & - & - & - & 47.4 & - & - \\
    % SimBaseline~\citep{xu2022simple} & COCO S. & - & - & 15.3 & - & 74.5 & - & - \\
    % ZegFormer~\citep{ding2022decoupling} & COCO S. & - & - & 16.4 & - & 73.3 & - & - \\
    LSeg~\citep{li2022languagedriven} & COCO S. & 3.8 & 7.8 & 18.0 & 46.5 & - & - & - \\
    OVSeg~\citep{liang2023open} & COCO S. & 9.0 & 12.4 & 29.6 & 55.7 & - & 94.5 & - \\
    SAN~\citep{xu2023side} & COCO S. & 13.7 & 17.1 & 33.3 & 60.2 & - & 95.5 & - \\
    OpenSeg~\citep{ghiasi2022scaling} & COCO P. + C. & 6.3 & 9.0 & 21.1 & 42.1 & - & - & - \\
    % ODISE$^{*}$~\citep{xu2023open} & COCO P. + C. & 11.0 & 13.8 & 28.7 & 55.3 & 82.7 & - & - \\
    ODISE~\citep{xu2023open} & COCO P. & 11.1 & 14.5 & 29.9 & 57.3 & 84.6 & - & - \\
    X-Decoder~\citep{zou2023generalized} & COCO P. + C. & - & - & 25.0 & - & - & - & 47.3 \\
    % OpenSEED~\citep{zhang2023simple} & COCO P. + O365 & - & - & 22.9 & - & - & - & 46.1 \\
    MaskCLIP~\citep{ding2023open} & COCO P. & 8.2 & 10.0 & 23.7 & 45.9 & - & - & - \\
    FC-CLIP~\citep{yu2023convolutions} & COCO P. & 14.8 & 18.2 & 34.1 & 58.4 & 81.8 & 95.4 & 56.2 \\
    \hline
    % \multicolumn{9}{l}{\textit{\textbf{Training-free method}}} \\
    % \hline
    % MaskCLIP & - & - & - & 11.9 & 26.4 & 43.4 & 74.9 & 24.9 \\
    % SCLIP & - & - & - & 16.1 & 34.2 & 59.1 & 80.4 & 32.2 \\
    % CaR & - & - & - & - & 30.5 & 67.6 & - & - \\
    % \hline
    \multicolumn{9}{l}{\textit{\textbf{Weakly-supervised method (image-text pair or image-text \& image-mask pair)}}} \\
    \hline
    GroupViT~\citep{xu2022groupvit} & GCC + YFCC & 4.3 & 4.9 & 10.4 & 23.4 & 52.3 & 79.7 & 18.5 \\
    ReCo~\cite{shin2022reco} & ImageNet-1K & - & - & 11.2 & 22.3 & 25.1 & 57.7 & 21.6 \\
    TCL~\citep{cha2023learning} & GCC & - & - & 17.1 & 33.9 & 55.0 & 83.2 & 24.3 \\
    OVSeg~\citep{xu2023learning} & CC4M & - & - & 5.6 & - & 53.8 & - & - \\
    SegCLIP~\citep{luo2023segclip} & CC3M + COCO C. & - & - & 8.7 & - & 52.6 & - & - \\
    CLIPpy~\citep{ranasinghe2023perceptual} & HQITP-134M & - & - & 13.5 & - & 52.2 & - & - \\
    MixReorg~\citep{cai2023mixreorg} & CC12M & - & - & 10.1 & 25.4 & 50.5 & - & - \\
    SAM-CLIP~\citep{wang2023sam} & M. 41M & - & - & 17.1 & 29.2 & 60.6 & - & - \\
    % \hline
    \rowcolor{lightblue}
    Baseline (Ours) & 50\% SA1B & 12.1 & 13.3 & 27.0 & 40.6 & 60.3 & 90.6 & 27.2 \\
    \rowcolor{lightblue}
    Unpair-Seg (Ours) & 50\% SA1B + M. 130K & \textbf{14.6} & \textbf{19.5} & \textbf{32.6} & \textbf{52.2} & \textbf{76.1} & \textbf{93.0} & \textbf{45.2} \\
    \hline
    \bottomrule
    \end{tabular}}
    \vspace{-0.5cm}
\end{table}
  
\myPara{Training configuration.}
We adopt our visual and text encoders as our ConvNext-Large CLIP model~\cite{liu2022convnet,radford2021learning} from OpenCLIP~\cite{ilharco2021openclip}.
For mask segmentation training, we mainly follow~\cite {cheng2022masked} and adopt a similar training recipe and losses without any particular design.
The input size of the image is $1024 \times 1024$, and we set the batch size to 64.
The model is optimised with AdamW~\cite{kingma2014adam,loshchilov2017decoupled}, where the learning rate is set to $1\times 10^{-4}$ and the weight decay is set to $5\times 10^{-2}$.
We train the model for 200k iterations and update the learning rate via the multi-step decay schedule. The parameters $\lambda_{bce}$, $\lambda_{dice}$, and $\lambda_{iou}$ are set to 2, 1, and 5, respectively.
Afterwards, during the training process of mask-entity alignment, we adopt a $20 \times 20$ point grid as the input visual prompt. The batch size and learning rate are set to 32 and $2\times 10^{-5}$, respectively. We train this stage for 30k iterations.
We conduct all training on 8 NVIDIA H100 80GB, 109 hours for the first stage, and 7 hours for the second stage.

\myPara{Evaluation \& metrics.}
We evaluate our model mainly on four tasks: open-vocabulary semantic segmentation, open-vocabulary panoptic segmentation, point-prompt segmentation, and box-prompt segmentation.
Following previous work~\cite{yu2023convolutions}, we adopt prompt engineering from~\cite{ghiasi2022scaling,xu2023open} and prompt templates from~\cite{gu2021open,li2022panoptic}.
For open-vocabulary semantic segmentation, we zero-shot evaluate the model on the COCO~\cite{lin2014microsoft}, ADE20K~\cite{zhou2017scene}, and PASCAL~\cite{everingham2010pascal} datasets.
Following previous work~\cite{yu2023convolutions}, we adopt the prompt engineering from~\cite{ghiasi2022scaling,xu2023open,gu2021open,li2022panoptic}.
The open-vocabulary semantic segmentation results are evaluated with the mean Intersection-over-Union (mIoU).
We evaluate the model on the COCO, ADE20K, and Cityscapes~\cite{cordts2016cityscapes} datasets for open-vocabulary panoptic segmentation.
We report the panoptic quality (PQ), semantic quality (SQ), and recognition quality (RQ) for open-vocabulary panoptic segmentation.
For point-prompt and box-prompt segmentation, from the oracle perspective, we report the 1-pt IoU and 1-box IoU on a wide range of datasets.
``Oracle'' denotes that we select the output mask with the max IoU by calculating the IoU between the prediction and ground-truth mask.
More details can be found in Appendix~\ref{sec:detail} and Appendix~\ref{sec:p_seg}.

\subsection{Main results}
\myPara{Open-vocabulary semantic segmentation.}
To showcase the significant reduction in the performance gap between fully-supervised and weakly-supervised methods, we conducted a comprehensive comparison across various datasets, including ADE20K (encompassing both 150 and 847 class variants)~\cite{zhou2017scene}, PASCAL Context (459 and 59 class variants)~\cite{everingham2010pascal}, PASCAL VOC (with 20 and 21 class categories)~\cite{everingham2010pascal}, and Cityscapes~\cite{cordts2016cityscapes}. As shown in Table~\ref{tab:ov_semseg}, Unpair-Seg outperforms weakly-supervised methods across different datasets by utilising independent image-mask and image-text pairs. Notably, Unpair-Seg performs impressive on the challenging PASCAL Context-459 and ADE20K-847 datasets, surpassing the state-of-the-art fully-supervised method FC-CLIP~\cite{yu2023convolutions}. Upon further analysis, we observed that FC-CLIP achieves significantly higher accuracy for in-vocabulary classes (corresponding to the COCO dataset) than out-of-vocabulary ones on the PASCAL Context-459 and ADE20K-847 datasets. More details can be found in Appendix~\ref{sec:in_out_iou}. Leveraging the diversity of image-text pairs, our method demonstrates superior capability in categorising general semantic classes.

\begin{table}[h]
    \centering
    \caption{
        \textbf{Open-vocabulary panoptic segmentation performance.}
        We mainly compare with the fully-supervised and unsupervised methods.
        ``COCO P.'', ``COCO'' and ``IN 1K'' denote the COCO panoptic, COCO image and ImageNet-1K datasets, respectively.
        We report PQ, SQ and RQ for all datasets.
        Fully-supervised method trained on the ``COCO P.'' datasets, so we show the results in grey. 
    }
    % \vspace{-0.2cm}
    \label{tab:ov_panoseg}
    \setlength{\tabcolsep}{1.3mm}{
    \footnotesize
    \begin{tabular}{l|c|ccc|ccc|ccc}
    \toprule
    \hline
     &  & \multicolumn{3}{c|}{\textbf{COCO}} & \multicolumn{3}{c|}{\textbf{ADE20K}} & \multicolumn{3}{c}{\textbf{Cityscapes}} \\
     \cline{3-11}
     \multirow{-2}{*}{\textbf{Method}} & \multirow{-2}{*}{\textbf{Training Data}} & \textbf{PQ} & \textbf{SQ} & \textbf{RQ} & \textbf{PQ} & \textbf{SQ} & \textbf{RQ} & \textbf{PQ} & \textbf{SQ} & \textbf{RQ} \\
    \hline
    \multicolumn{11}{l}{\textit{\textbf{Fully-supervised method (image-text-mask pair)}}} \\
    \hline
    MaskCLIP~\citep{ding2023open} & COCO P. & \textcolor{gray}{30.9} & \textcolor{gray}{-} & \textcolor{gray}{-} & 15.1 & 70.5 & 19.2 & - & - & - \\
    ODISE~\citep{xu2023open} & COCO P. & \textcolor{gray}{55.4} & \textcolor{gray}{-} & \textcolor{gray}{-} & 22.6 & - & - & 23.9 & 75.3 & 29.0 \\
    FC-CLIP~\citep{yu2023convolutions} & COCO P. & \textcolor{gray}{54.4} & \textcolor{gray}{83.0} & \textcolor{gray}{64.8} & 26.8 & 71.6 & 32.3 & 44.0 & 75.4 & 53.6 \\
    OPSNet~\citep{chen2023open} & COCO P. & \textcolor{gray}{57.9} & \textcolor{gray}{84.1} & \textcolor{gray}{68.2} & 19.0 & 52.4 & 23.0 & 41.5 & 67.5 & 50.0 \\
    \hline
    \multicolumn{11}{l}{\textit{\textbf{Unsupervised method (Unlabelled image)}}} \\
    \hline
    CutLER+STEGO~\citep{wang2023cut} & IN 1K + COCO & 12.4 & 64.9 & 15.5 & - & - & - & 12.4 & 36.1 & 15.2 \\
    U2Seg~\citep{niu2023unsupervised} & IN 1K + COCO & 16.1 & 71.1 & 19.9 & - & - & - & 17.6 & 52.7 & 21.7 \\
    \hline
    \multicolumn{11}{l}{\textit{\textbf{Weakly-supervised method (image-text \& image-mask pair)}}} \\
    \hline
    \rowcolor{lightblue}
    Baseline (Ours) & 50\% SA1B & 24.4 & 79.6 & 30.0 & 19.1 & 78.1 & 23.7 & 17.3 & 71.4 & 22.0 \\
    \rowcolor{lightblue}
    Unpair-Seg (Ours) & 50\% SA1B + M. 130K & \textbf{28.8} & \textbf{80.0} & \textbf{35.2} & \textbf{22.6} & \textbf{74.0} & \textbf{28.0} & \textbf{26.8} & \textbf{70.9} & \textbf{33.4} \\
    \hline
    \bottomrule
    \end{tabular}}
    \vspace{-0.4cm}
\end{table}

\begin{table}[h]
    \centering
    \centering
    \caption{
        \textbf{Promptable segmentation performance.}
        We compare our method with SAM~\cite{kirillov2023segment}. Given point or box prompts, we select the output masks with max IoU by calculating the IoU with the ground-truth masks. We report 1-pt IoU on the COCO and ADE20K panoptic segmentation, and 1-box IoU on the COCO and ADE20K instance segmentation datasets, respectively.
    }
    % \vspace{-0.2cm}
    \label{tab:prompt_seg}
    \setlength{\tabcolsep}{1.1mm}{
    \footnotesize
    \begin{tabular}{l|c|c|c|c|c|c}
    \toprule
    \hline
    \multirow{2}{*}{\textbf{Method}} &  \multirow{2}{*}{\textbf{Param}} & \multirow{2}{*}{\textbf{Training Data}} & \multicolumn{2}{c|}{\textbf{COCO}} & \multicolumn{2}{c}{\textbf{ADE20K}} \\
    \cline{4-7}
     ~ & ~ & ~ & 1-pt IoU (\%) & 1-box IoU (\%) & 1-pt IoU (\%) & 1-box IoU (\%) \\
    \hline
    SAM-Base & 93M & 100\% SA-1B & 42.7 & 72.8 & 41.6 & 74.7 \\
    SAM-Large & 312M & 100\% SA-1B & 64.3 & 77.9 & 61.1 & 77.7 \\
    SAM-Huge & 641M & 100\% SA-1B & 66.7 & \textbf{78.0} & 63.3 & \textbf{78.3} \\
    \rowcolor{lightblue}
    Unpair-Seg (Ours) & 221M & 50\% SA-1B & \textbf{80.8} & 76.9 & \textbf{76.9} & 77.3 \\
    \hline
    \bottomrule
    \end{tabular}}
    \vspace{-0.4cm}
\end{table}

\myPara{Open-vocabulary panoptic segmentation.}
Existing weakly-supervised methods only utilise text supervision, which makes it challenging to distinguish different instances with the same semantic class. To the best of our knowledge, Unpair-Seg is the first attempt to tackle the difficult task of open-vocabulary panoptic segmentation with weak supervision. Due to the absence of benchmarks, we evaluate our method on the COCO, ADE20K, and Cityscapes datasets and compared it comprehensively with fully-supervised and unsupervised methods in Table~\ref{tab:ov_panoseg}.
Compared to unsupervised methods, our approach outperforms U2Seg~\cite{niu2023unsupervised} by 12.7\% PQ, 8.9\% SQ, and 15.3\% RQ on the COCO panoptic segmentation task. Even more impressively, Unpair-Seg exceeds specific fully-supervised methods. For instance, we achieve a noticeable gain of 3.6\% PQ than OPSNet~\cite{chen2023open} on the ADE20K dataset and a remarkable improvement of 2.9\% PQ than ODISE~\cite{xu2023open} on the Cityscapes panoptic segmentation.
These results show that our method effectively captures fine-grained spatial relations through weak supervision, thereby enhancing the practical usefulness of weakly-supervised open-vocabulary segmentation.

\myPara{Promptable segmentation.}
Following SAM~\cite{kirillov2023segment}, we evaluate segmenting an object from one input point, which is challenging as one point can refer to objects at multiple levels of detail.
To provide visual cues, we have implemented a uniform $20 \times 20$ point grid as the interactive point prompt and use the actual bounding boxes as box prompts.
We compare our method with three variants of SAM, as shown in Table~\ref{tab:prompt_seg}.
By using the proposed many-to-many matching, our method can predict multi-granularity masks when employing point prompts. In particular, Unpair-Seg significantly outperforms SAM-Large in terms of 1-point IoU by 16.5\% and 15.8\% on the COCO and ADE20K panoptic segmentation datasets, respectively.
We also show some visualisations in Figure~\ref{fig:vis} and more experimental results and visualisations can be found in Appendix~\ref{sec:p_seg} and \ref{sec:vis_p}.

\begin{table}[t]
    \centering
    \begin{minipage}[t]{0.45\textwidth}
        \centering
        \caption{
            \textbf{Vision-language large model.}
            ``\textit{LLaVA.}'' and ``\textit{InternLM-VL.}'' denote the LLaVA-v1.5-7B and InternLM-XComposer2-VL-7B models.
            % We report PQ and mIoU on the COCO and PASCAL Context-59 datasets, respectively.
        }
        % \vspace{-0.2cm}
        \label{tab:ablation1}
        \setlength{\tabcolsep}{4.2mm}{
        \scriptsize
        \begin{tabular}{c|cc}
        \toprule
        \hline
        \multirow{2}{*}{\textbf{Method}} & \textbf{COCO} & \textbf{PC-59} \\
        \cline{2-3}
         ~ & PQ (\%) & mIoU (\%) \\
        \hline
        \textit{LLaVA.} & 25.4 & 42.6 \\
        \textit{InternLM-VL.} & \textbf{28.8} & \textbf{52.2} \\
        \hline
        \bottomrule
        \end{tabular}}
        % \vspace{-0.2cm}
    \end{minipage}
    \hfill
    \begin{minipage}[t]{0.48\textwidth}
        \centering
        \caption{
            \textbf{Design choice of the feature adapter.}
            ``\textit{Cross Attn.}'' and ``\textit{Low Rank.}'' denote the cross attention-based and low rank-base feature adapters.
            % We report PQ and mIoU on the COCO and PASCAL Context-59 datasets, respectively.
        }
        % \vspace{-0.2cm}
        \label{tab:ablation2}
        \setlength{\tabcolsep}{4.2mm}{
        \scriptsize
        \begin{tabular}{c|cc}
        \toprule
        \hline
        \multirow{2}{*}{\textbf{Method}} & \textbf{COCO} & \textbf{PC-59} \\
        \cline{2-3}
         ~ & PQ (\%) & mIoU (\%) \\
        \hline
        \textit{Cross Attn.} & 27.1 & 47.4 \\
        \textit{Low Rank.} & \textbf{28.8} & \textbf{52.2} \\
        \hline
        \bottomrule
        \end{tabular}}
        % \vspace{-0.2cm}
    \end{minipage}
\vspace{-0.4cm}
\end{table}
\begin{table}[t]
    \centering
    \begin{minipage}[t]{0.45\textwidth}
        \centering
        \caption{
            \textbf{Mask-entity matching strategy.}
            ``\textit{Greedy.}'' and ``\textit{Bipartite.}'' denote the confidence greedy and bipartite match strategies.
            % We report PQ and mIoU on the COCO and PASCAL Context-59 datasets, respectively.
        }
        % \vspace{-0.2cm}
        \label{tab:ablation3}
        \setlength{\tabcolsep}{3.5mm}{
        \scriptsize
        \begin{tabular}{c|cc}
        \toprule
        \hline
        \multirow{2}{*}{\textbf{Method}} & \textbf{COCO} & \textbf{PC-59} \\
        \cline{2-3}
         ~ & PQ (\%) & mIoU (\%) \\
        \hline
        \textit{Greedy. (1 sigma)} & 26.1 & 43.7 \\
        \textit{Greedy. (2 sigma)} & 26.0 & 43.6 \\
        \textit{Bipartite.} & \textbf{28.8} & \textbf{52.2} \\
        \hline
        \bottomrule
        \end{tabular}}
        % \vspace{-0.6cm}
    \end{minipage}
    \hfill
    \begin{minipage}[t]{0.48\textwidth}
        \centering
        \caption{
            \textbf{Multi-scale ensemble.}
            We ensemble multi-scale visual embeddings extracted by the CLIP model to stablise the matching process.
            % We report PQ and mIoU on the COCO and PASCAL Context-59 datasets, respectively.
        }
        % \vspace{-0.2cm}
        \label{tab:ablation4}
        \setlength{\tabcolsep}{3.5mm}{
        \scriptsize
        \begin{tabular}{c|cc}
        \toprule
        \hline
        \multirow{2}{*}{\textbf{Size}} & \textbf{COCO} & \textbf{PC-59} \\
        \cline{2-3}
         ~ & PQ (\%) & mIoU (\%) \\
        \hline
        896 & 27.4 & 49.0 \\
        896, 1024 & 28.2 & 50.0 \\
        896, 1024, 1152 & \textbf{28.8} & \textbf{52.2} \\
        \hline
        \bottomrule
        \end{tabular}}
        % \vspace{-0.6cm}
    \end{minipage}
\vspace{-0.4cm}
\end{table}

\begin{figure}[!ht]
    \centering
    \includegraphics[width=\textwidth]{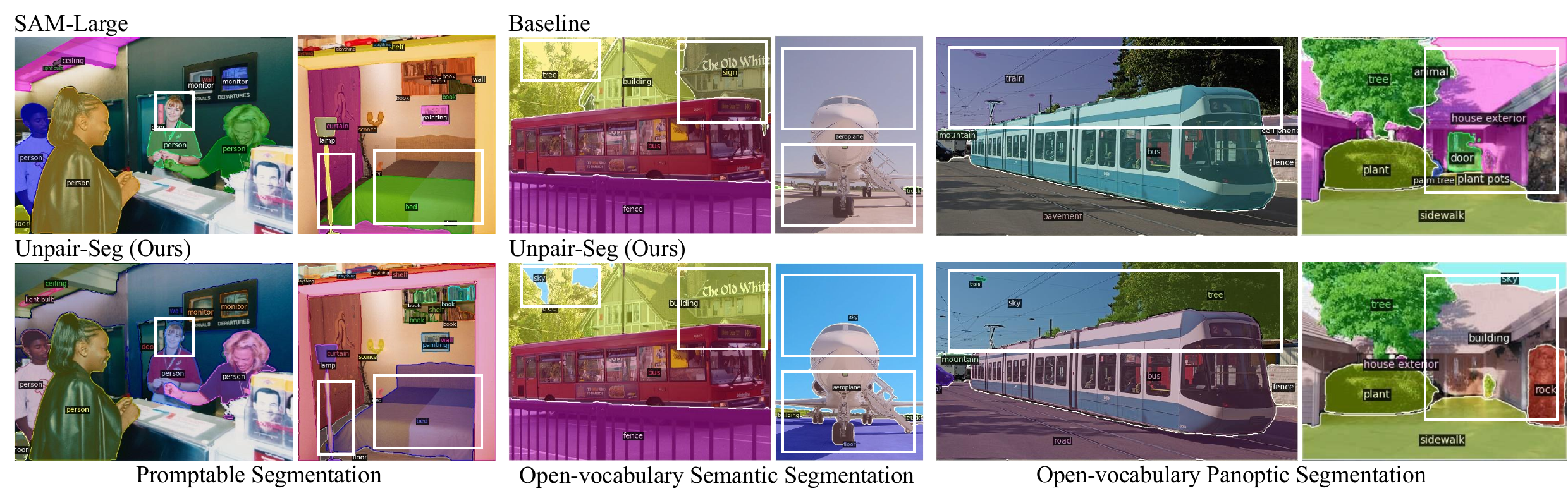}
    % \vspace{-0.2cm}
    \caption{\textbf{Visualisaton.} We show prediction results on three tasks: promptable, open-vocabulary semantic, and open-vocabulary panoptic segmentation. The results are best viewed in color.}
    \label{fig:vis}
\vspace{-0.6cm}
\end{figure}

\subsection{Ablation study}
We conducted a thorough ablation study to assess the impact of the main components in our framework. Compared to the simple baseline, Unpair-Seg achieved significant improvements of 2.5\%, 6.2\%, and 18.0\% mIoU on the ADE20k-847, PASCAL Context-459, and Cityscapes semantic segmentation, and 4.4\% and 3.5\% PQ on the COCO and ADE20K panoptic segmentation, as shown in Table~\ref{tab:ov_semseg} and Table~\ref{tab:ov_panoseg}. These results demonstrate the effectiveness of our method in aligning objects in images and entities in text descriptions, generalising the CLIP embedding space from the image to the pixel level. Qualitatively, we illustrate a visualisation comparison between the baseline and our proposed Unpair-Seg in Figure~\ref{fig:vis}.
Furthermore, we explored the impact of different vision-language large models on the quality of image-text pairs. Our ablation study on two models, LLaVA~\cite{liu2023visual} and InternLM-VL~\cite{internlmxcomposer2}, detailed in Table~\ref{tab:ablation1}, revealed their influence on model performance. High-quality captions were found to help the model learn accurate mask-entity alignment.
During the mask-entity alignment, we use a feature adapter to enhance regional embeddings, followed by aligning with text embedding. In Table~\ref{tab:ablation2}, we design two alternative modules to aggregate multi-scale visual information and improve the quality of regional embedding. A low-rank feature adapter that is adopted in our framework achieves better results.
As illustrated in Table~\ref{tab:ablation3}, we design two mask-entity matching strategies. The greedy method assigns an entity for each predicted mask based on the maximum confidence. The Bipartite method optimises an assignment plan to achieve the global minimum cost, which obtains better results than the greedy method.
% Finally, considering the variety of object scales, we employ a multi-scale ensemble to improve the quality of CLIP visual embeddings, obtaining confident mask-entity pairs. We adopt the sizes of $869\times 896$, $1024 \times 1024$, and $1152 \times 1152$ as default.
Finally, considering the variety of object scales, we employ a multi-scale ensemble that extracts CLIP visual embeddings from multiple resolution images and ensemble them into one embedding, which is used to obtain confident mask-entity pairs. As shown in Table~\ref{tab:ablation4}, we adopt the sizes of $869\times 896$, $1024 \times 1024$, and $1152 \times 1152$ as default.
\section{Conclusion}
\label{sec:conclusion}
This paper introduces Unpair-Seg, a novel framework for weakly-supervised open-vocabulary segmentation. The key strength of Unpair-Seg lies in its use of unpaired mask-text supervision, which reduces the need for labor-intensive annotations. To mitigate inherent noise in the mask-entity matching process, we employ a vision-language large model for precise entity extraction and a multi-scale matching strategy for stable mask-entity alignment. By predicting binary masks and generating pseudo labels through confident mask-text entity pairs, Unpair-Seg significantly outperforms previous weakly-supervised methods across multiple segmentation tasks and datasets. This significant improvement brings us closer to the performance of fully-supervised methods in real-world scenarios. However, it's important to acknowledge that the framework's performance may be influenced by the quality of image-text pairs and initial mask predictions in complex or ambiguous scenes. These aspects will be the focus of our future work.
\bibliography{icml2024}
\bibliographystyle{icml2024}
%%%%%%%%%%%%%%%%%%%%%%%%%%%%%%%%%%%%%%%%%%%%%%%%%%%%%%%%%%%%%%%%%%%%%%%%%%%%%%%
%%%%%%%%%%%%%%%%%%%%%%%%%%%%%%%%%%%%%%%%%%%%%%%%%%%%%%%%%%%%%%%%%%%%%%%%%%%%%%%
% APPENDIX
%%%%%%%%%%%%%%%%%%%%%%%%%%%%%%%%%%%%%%%%%%%%%%%%%%%%%%%%%%%%%%%%%%%%%%%%%%%%%%%
%%%%%%%%%%%%%%%%%%%%%%%%%%%%%%%%%%%%%%%%%%%%%%%%%%%%%%%%%%%%%%%%%%%%%%%%%%%%%%%
\newpage
\appendix
\onecolumn
% \section{You \emph{can} have an appendix here.}

% You can have as much text here as you want. The main body must be at most $8$ pages long.
% For the final version, one more page can be added.
% If you want, you can use an appendix like this one.  

% The $\mathtt{\backslash onecolumn}$ command above can be kept in place if you prefer a one-column appendix, or can be removed if you prefer a two-column appendix.  Apart from this possible change, the style (font size, spacing, margins, page numbering, etc.) should be kept the same as the main body.
\section{Framework details}
\label{sec:detail}

\myPara{Inputs.}
For input images, we initially apply a random horizontal flip to each image. Subsequently, the image is randomly scaled to a resolution within the range of $716 \times 716$ to $1075 \times 1075$ resolutions. Finally, a crop of $896 \times 896$ resolutions is extracted from the scaled image to serve as the input. Regarding the category names, we construct a sentence using a prompt and tokenise it using a lower-cased byte pair encoding (BPE). For the visual prompt, we create a uniform grid of points with dimensions $h \times w$, which is aligned with the center of the pixels.

\myPara{CLIP encoder.}
In general, the CLIP encoder can be any architecture.
Motivated by the scalability of different input resolutions, we employ a ConvNext-based CLIP model to serve as both the image and text encoders. The image encoder is configured as a ConvNext-Large model, comprising four stages. Each stage contains a different number of blocks: 3 in the first, 3 in the second, 27 in the third, and 3 in the fourth.
In contrast, the text encoder is structured as a 16-layer transformer, each layer being 768 units wide and featuring 12 attention heads. We harness the power of multi-scale features extracted by the image encoder. These features are represented as feature maps of varying widths and scales: a 192-wide feature map downscaled by a factor of 4, a 384-wide map downscaled by 8, a 768-wide map downscaled by 16, and a 1536-wide map downscaled by 32.

\myPara{Pixel decoder.}
Following Mask2Former~\cite{cheng2022masked}, we incorporate a lightweight pixel decoder based on the widely used Feature Pyramid Network (FPN) architecture~\cite{lin2017feature}. We first adopt a 6-layer multi-scale deformable transformer on the multi-scale features to aggregate contextual information. Afterward, we upscale the low-resolution feature map in the decoder by a factor of 2 and then combine it with the corresponding resolution feature map from the backbone, which has been projected to match channel dimensions. This projection is achieved through a 1$\times$1 convolution layer, followed by Group Normalization (GroupNorm). Subsequently, the merged features undergo further processing with an additional 3$\times$3 convolution layer, complemented by GroupNorm and ReLU activation. This procedure is applied iteratively, beginning with the 32$\times$ downscaled feature map, until we attain a final feature map that is 4$\times$ downscaled. To generate pixel-wise embeddings, a single 1$\times$1 convolution layer is applied at the end. Throughout the pixel decoder, all feature maps maintain a consistent dimension of 256 channels.

\myPara{Visual prompt encoder.}
We have two types of visual prompts, including points and bounding boxes.
Each type of prompt is mapped to 256-wide embeddings as follows.
A point is first converted into a small bounding box.
We represent a bounding box by an embedding pair, including the sine position embedding of its top-left and bottom-right corners.
Afterward, we use two extra learnable embeddings to indicate these two corners.

\begin{figure}[h]
    \begin{center}
    \centerline{
        \includegraphics[width=0.5\textwidth]{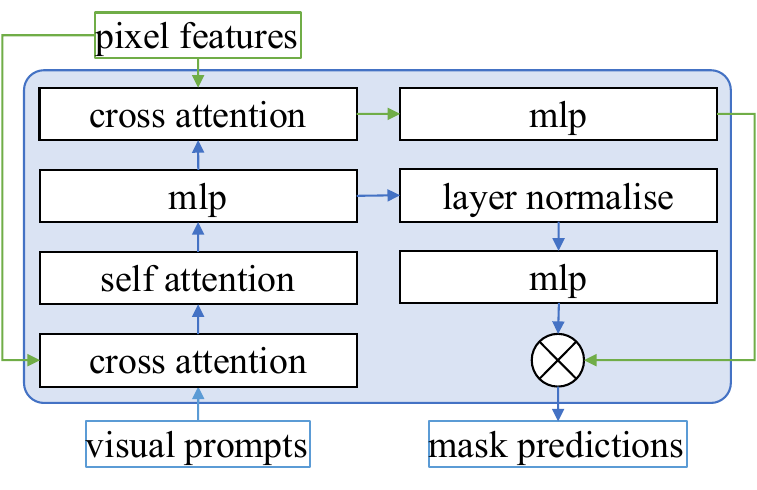}
    }
    \caption{
        \textbf{Architecture of the mask decoder layer.}
        This decoder layer updates both visual prompt embeddings and pixel features by the cross-attention layers.
        The self-attention layer is used to update visual prompts.
        At each attention layer, positional encodings are added to the pixel features, and the entire original visual prompts (including position encoding) are added to the updated visual prompts.
    }
    \label{fig:decoder}
    \end{center}
\end{figure}
\myPara{Mask decoder.}
Following Mask2Former, we use a similar transformer decoder design. Each visual prompt embedding within our framework is coupled with a sine positional embedding. As depicted in Figure~\ref{fig:decoder}, our approach utilizes six mask decoder layers, and we apply the same loss function after each layer.
Moreover, every decoder layer includes a cross-attention layer. This layer is crucial as it ensures that the final pixel features are enriched with essential geometric information, such as point coordinates and bounding boxes. Besides, in each decoder layer, we use a cross-attention layer, which ensures that the final pixel features have access to critical geometric information (\eg, point coordinates, and boxes).
Finally, visual prompt embeddings are fed to a layer normalization, followed by processing through a multiple perception layer (MLP). These processed embeddings are then dot-producted with the pixel features, culminating in the generation of mask predictions.

\begin{figure}[h]
    \begin{center}
    \centerline{
        \includegraphics[width=0.5\textwidth]{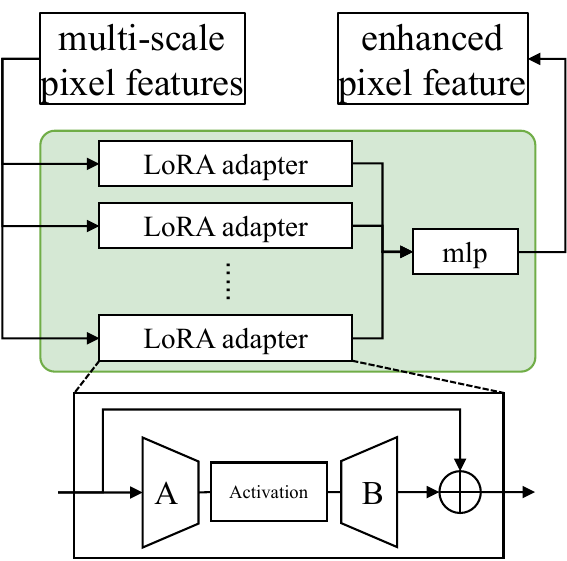}
    }
    \caption{
        \textbf{Architecture of the multi-scale feature adapter.}
        Each low-rank feature adapter is used to update specific downscaled features.
    }
    \label{fig:adapter}
    \end{center}
\end{figure}
\myPara{Multi-scale feature adapter.}
The multi-scale feature adapter illustrated in Figure~\ref{fig:adapter} is designed to refine and enhance pixel features at various scales. This adapter is composed of multiple Low-rank Adapters~\cite{hu2021lora}, each specifically tailored to update features at a particular scale. Each feature adapter incorporates two linear layers, denoted as A and B, with a non-linear activation function placed in between. The linear layers A and B serve to transform the input data linearly, while the activation function introduces non-linearity, allowing for the modeling of more complex relationships in the data. Each adapter is responsible for handling features at a specific scale, suggesting a hierarchical approach to feature refinement. This modular design allows for focused and specialized treatment of features depending on their resolution and semantic complexity, which can be particularly advantageous for dense prediction tasks.

\begin{figure*}
    \centering
    \vspace{-0.2cm}
    \includegraphics[width=\textwidth]{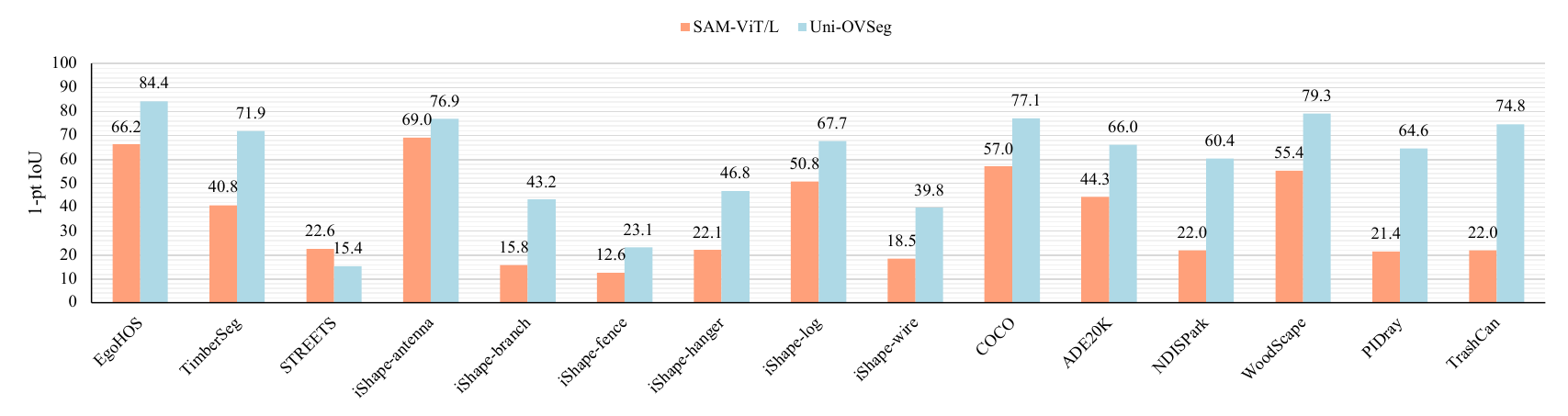}
    \vspace{-0.4cm}
    \caption{\textbf{Point-prompt segmentation performance.} We compare our method with SAM-ViT/L~\cite{kirillov2023segment}. Given a $20\times 20$ point grid as visual prompt, we select the output masks with max IoU by calculating the IoU with the ground-truth masks. We report 1-pt IoU for all datasets.}
    \label{fig:point_seg1}
    \vspace{-0.4cm}
\end{figure*}
\begin{figure*}[!t]
    \centering
    \vspace{-0.2cm}
    \includegraphics[width=\textwidth]{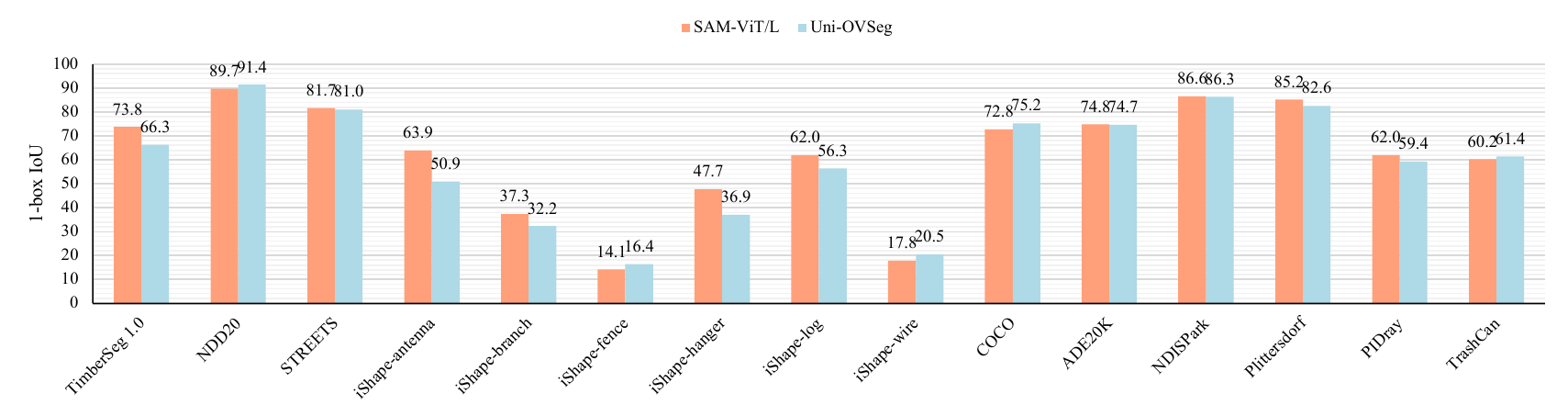}
    \vspace{-0.2cm}
    \caption{\textbf{Box-prompt segmentation performance.} We compare our method with SAM-ViT/L~\cite{kirillov2023segment}. Given a ground-truth box as the visual prompt, we select the output masks with max IoU by calculating the IoU with the ground-truth masks. We report 1-pt IoU for all datasets.}
    \label{fig:box_seg1}
    \vspace{-0.2cm}
\end{figure*}

\begin{figure*}[!t]
    \centering
    \vspace{-0.2cm}
    \includegraphics[width=\textwidth]{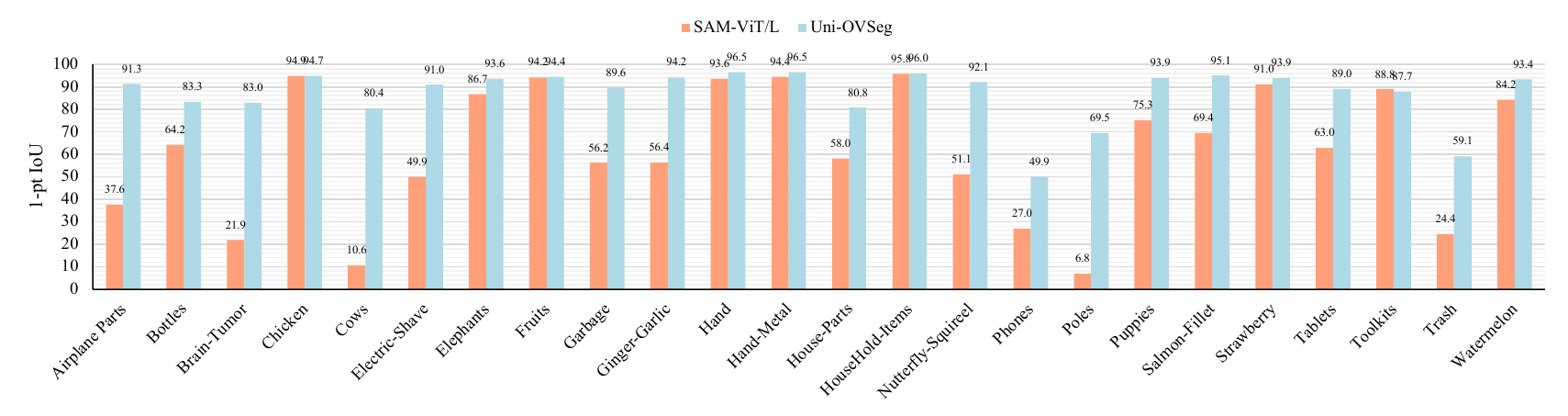}
    \vspace{-0.2cm}
    \caption{\textbf{Point-prompt segmentation performance on the SegInW dataset.} We compare our method with SAM-ViT/L~\cite{kirillov2023segment}. Given a $20\times 20$ point grid as a visual prompt, we select the output masks with max IoU by calculating the IoU with the ground-truth masks. We report 1-pt IoU for all datasets.}
    \label{fig:point_seg2}
    \vspace{-0.2cm}
\end{figure*}

\begin{figure*}[!t]
    \centering
    \vspace{-0.2cm}
    \includegraphics[width=\textwidth]{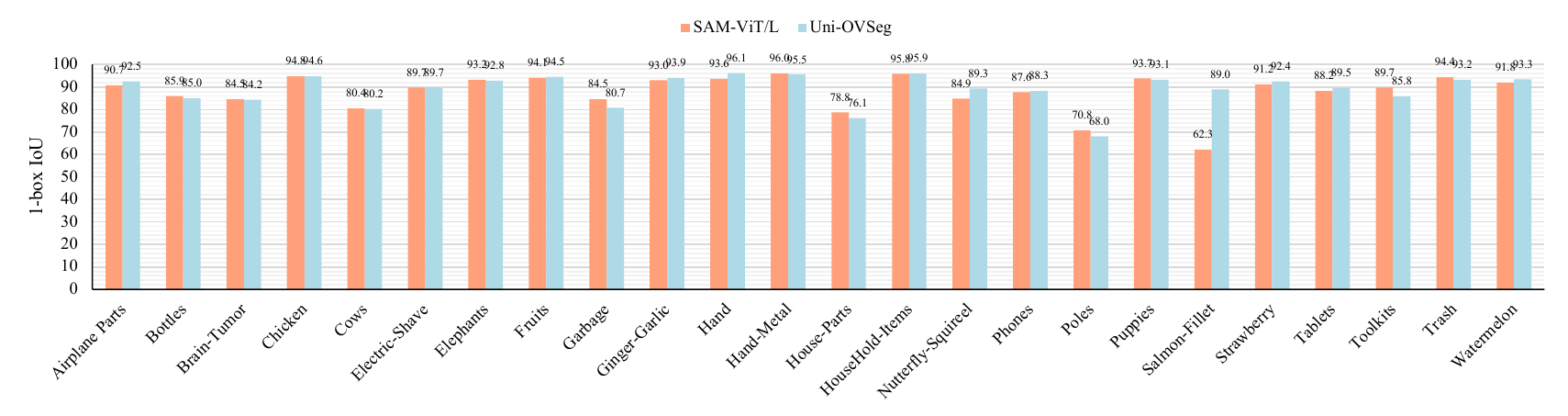}
    \vspace{-0.2cm}
    \caption{\textbf{Box-prompt segmentation performance on the SegInW dataset.} We compare our method with SAM-ViT/L~\cite{kirillov2023segment}. Given a ground-truth box as the visual prompt, we select the output masks with max IoU by calculating the IoU with the ground-truth masks. We report 1-box IoU for all datasets.}
    \label{fig:box_seg2}
    \vspace{-0.2cm}
\end{figure*}

\begin{table}[h]
    \centering
    \begin{minipage}[t]{0.48\textwidth}
        \centering
        \caption{
            \textbf{In- and out-vocabulary mIoU on ADE20K-847 semantic segmentation.}
            We report the results of FC-CLIP and our method, respectively.
        }
        % \vspace{-0.2cm}
        \label{tab:iou1}
        \setlength{\tabcolsep}{2mm}{
        \scriptsize
        \begin{tabular}{c|cc}
        \toprule
        \hline
        \textbf{Method} & \textbf{In.} (101 classes) & \textbf{Out.} (746 classes) \\
        \hline
        FC-CLIP~\cite{yu2023convolutions} & \textbf{35.1} & 12.0 \\
        Unpair-Seg & 32.5 & \textbf{12.1} \\
        \hline
        \bottomrule
        \end{tabular}}
        % \vspace{-0.2cm}
    \end{minipage}
    \hfill
    \begin{minipage}[t]{0.48\textwidth}
        \centering
        \caption{
            \textbf{In- and out-vocabulary mIoU on PASCAL Context-459 semantic segmentation.}
            We report the results of FC-CLIP and our method, respectively.
        }
        % \vspace{-0.2cm}
        \label{tab:iou2}
        \setlength{\tabcolsep}{2mm}{
        \scriptsize
        \begin{tabular}{c|cc}
        \toprule
        \hline
        \textbf{Method} & \textbf{In.} (85 classes) & \textbf{Out.} (374 classes) \\
        \hline
        FC-CLIP~\cite{yu2023convolutions} & \textbf{47.5} & 9.5 \\
        Unpair-Seg & 42.4 & \textbf{12.7} \\
        \hline
        \bottomrule
        \end{tabular}}
        % \vspace{-0.2cm}
    \end{minipage}
\vspace{-0.4cm}
\end{table}
\section{In- \& Out-vocabulary IoU comparison}
\label{sec:in_out_iou}
As shown in Table~\ref{tab:iou1} and Table~\ref{tab:iou2}, we report in-vocabulary and out-vocabulary IoU on ADE20K-847 and PASCAL Context-459 datasets. FC-CLIP, trained on the COCO panoptic dataset in a fully supervised manner, exhibits higher mIoU for in-vocabulary classes (35.1\% on ADE20K-847 and 47.5\% on PASCAL Context-847), suggesting a tendency to overfit on these classes, thus limiting its open-vocabulary segmentation ability. In contrast, Unpair-Seg, which leverages the text diversity from large amounts of image-text pairs, demonstrates competitive in-vocabulary performance (32.5\% on ADE20K-847 and 42.4\% on PASCAL Context-847, respectively) and significantly better out-of-vocabulary mIoU (12.1\% and 12.7\%), indicating superior generalisation to unseen classes and robustness in sophisticated scenarios.

\section{Promptable segmentation}
\label{sec:p_seg}
\myPara{Evaluation details.}
We perform a prompt segmentation evaluation on a wide range of datasets in various domains. For the point prompt, we adopt a uniform point grid $h \times w$ as input prompts (\eg, 20 $\times$ 20). For the box prompt, we use ground-truth bounding boxes as input prompts.
1-pt IoU denotes the oracle performance of one point by evaluating the intersection-over-union (IoU) of the predicted masks that best match ground truth.
1-box IoU denotes is similar to 1-pt IoU.
More evaluation results are reported in Figure~\ref{fig:point_seg1}, Figure~\ref{fig:box_seg1}, Figure~\ref{fig:point_seg2} and Figure~\ref{fig:box_seg2}.

\begin{table*}[t]
    \centering
    \caption{
        Segmentation datasets used to evaluate promptable segmentation with point and box prompts.
        The 11 datasets cover a broad range of domains, which are illustrated in ``image type''.
        % To make our evaluation efficient, we subsample datasets that have more than 15k masks.
        % Specifically, we randomly sampled images so that the total number of masks in the images is \app10k.
    }
    \label{tab:dataset}
    \setlength{\tabcolsep}{1mm}{
    \footnotesize
    \begin{tabular}{c|c|c|c}
    \toprule
    \hline
     Dataset & Image type & Mask type & Description \\
    \hline
    \makecell[c]{Egocentric Hand-Object \\Segmentation (EgoHOS)\\ \cite{zhang2022fine}} & Egocentric & Instance & \makecell[c]{Fine-grained egocentric hand-object \\segmentation dataset. Dataset contains mask\\ annotations for existing datasets.} \\
    \hline
    \makecell[c]{TimberSeg 1.0\\ (TimberSeg)\\ \cite{fortin2022instance}} & Logs & Instance & \makecell[c]{Segmentation masks of individual logs\\ in piles of timber in various environments\\ and conditions. Images are taken from\\ an operator’s point-of-view.} \\
    \hline
    \makecell[c]{STREETS\\ \cite{snyder2019streets}} & Traffic camera & Instance & \makecell[c]{Segmentation masks of cars\\ in traffic camera footage.} \\
    \hline
    \makecell[c]{iShape\\ \cite{yang2021ishape}} & Irregular shapes & Instance & \makecell[c]{Segmentation masks of irregular shapes\\ like antennas, logs, shapes fences,\\ and hangers.} \\
    \hline
    \makecell[c]{COCO\\ \cite{lin2014microsoft}} & Scenes & Instance & \makecell[c]{Segmentation masks of complex everyday \\scenes containing common objects in \\their natural context.} \\
    \hline
    \makecell[c]{ADE20K\\ \cite{zhou2017scene}} & Scenes & Instance & \makecell[c]{Object and part segmentation masks\\ for images from SUN and Places datasets.} \\
    \hline
    \makecell[c]{Night and Day Instance\\ Segmented Park \\(NDISPart)\\ \cite{ciampi2021domain}} & Parking lots & Instance & \makecell[c]{Images of parking lots from video footage\\ taken at day and night during different weather \\conditions and camera angles \\for vehicle segmentation.} \\
    \hline
    \makecell[c]{WoodScape \\ \cite{yogamani2019woodscape}} & Fisheye driving & Instance & \makecell[c]{Fisheye driving dataset with segmentation\\ masks. Images are driving taken from four\\ surround-view cameras.} \\
    \hline
    \makecell[c]{PIDray\\ \cite{zhang2023pidray}} & X-ray & Instance & \makecell[c]{Segmentation masks of prohibited items \\in X-ray images of baggage.} \\
    \hline
    \makecell[c]{TrashCan\\ \cite{hong2020trashcan}} & Underwater & Instance & \makecell[c]{Segmentation masks of trash in images\\ taken by underwater ROVs. Images are \\sourced from the J-EDI dataset.} \\
    \hline
    \makecell[c]{Segmentation in the wild\\ (SegInW)\\ \cite{zou2023generalized}} & Multiple domain & Instance & \makecell[c]{This dataset consists of 25 free \\public Segmentation datasets, \\crowd-sourced on roboflow.com.} \\
    \hline
    \bottomrule
    \end{tabular}}
\end{table*}

\myPara{Dataset details.}
A description of each dataset is given in Table~\ref{tab:dataset}.
The iShape dataset has 6 subsets, including antenna, branch, fence, hanger, log, and wire. 

\section{Broader impacts}
\label{sec:impact}
Our approach represents a significant advancement in open-vocabulary segmentation by reducing the reliance on labour-intensive image-mask-text triplet annotations. This innovation has the potential to make cutting-edge vision perception systems more accessible, offering significant benefits across various sectors, such as environmental monitoring and autonomous vehicles. Developing a more efficient and accurate vision perception system contributes to the broader field of AI, potentially leading to more innovative applications and research in machine learning, computer vision, and related areas. As with any AI model, the risk of bias in the data used for training is a crucial concern. Efforts must be made to ensure the datasets are diverse and representative to avoid perpetuating or amplifying biases.

\section{Visualisation}
\label{sec:vis_p}
We illustrate a wide range of visualisations of prompt segmentation and open-vocabulary segmentation across multiple datasets.

\begin{figure*}[h]
    \centering
    \vspace{-0.2cm}
    \includegraphics[width=\textwidth]{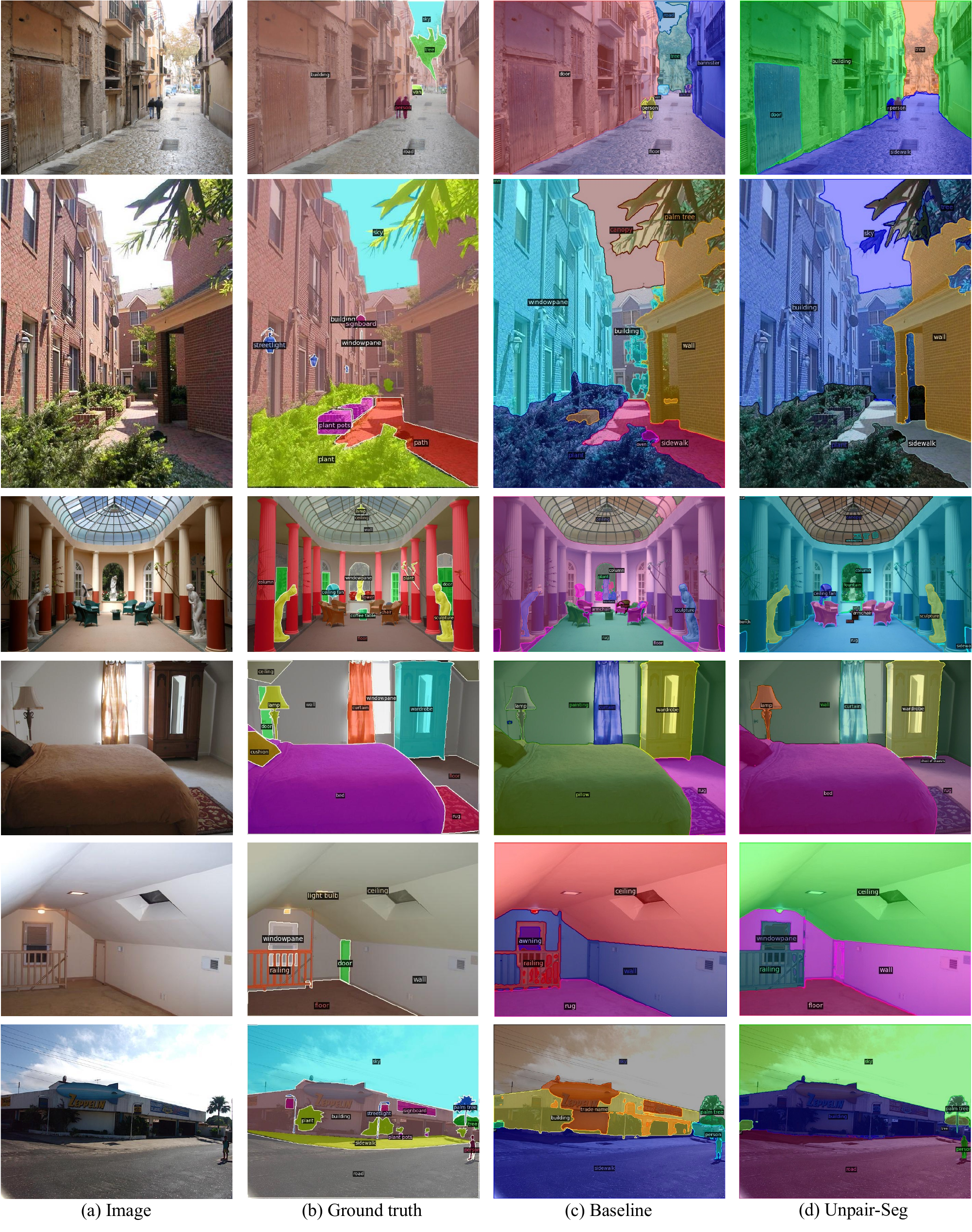}
    \vspace{-0.2cm}
    \caption{Visualisation of open-vocabulary segmentation between the baseline and Unpair-Seg.}
    \label{fig:p_sem_1}
    \vspace{-0.2cm}
\end{figure*}
\begin{figure*}[h]
    \centering
    \vspace{-0.2cm}
    \includegraphics[width=\textwidth]{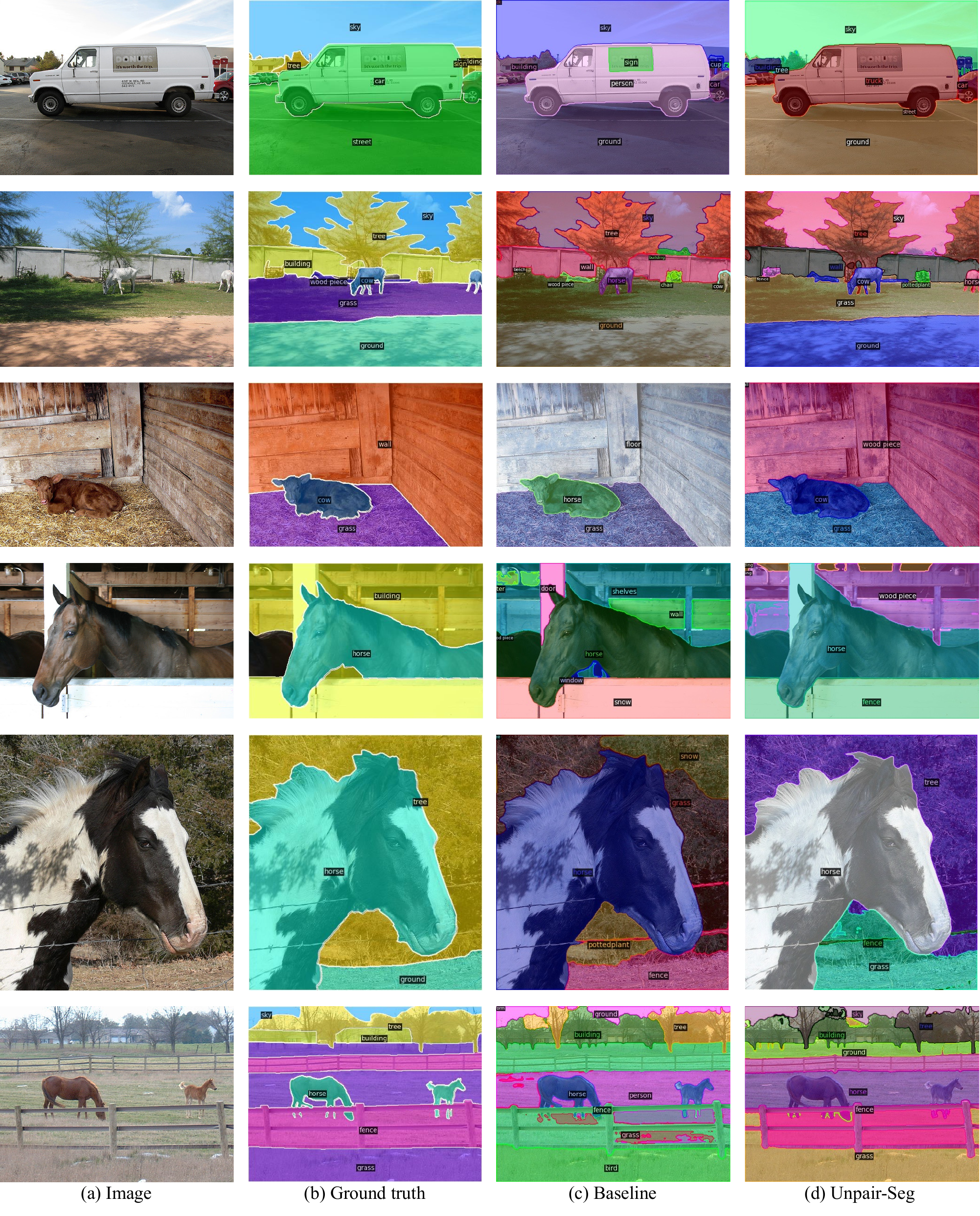}
    \vspace{-0.2cm}
    \caption{Visualisation of open-vocabulary segmentation between the baseline and Unpair-Seg.}
    \label{fig:p_sem_2}
    \vspace{-0.2cm}
\end{figure*}
\begin{figure*}[h]
    \centering
    \vspace{-0.2cm}
    \includegraphics[width=0.95\textwidth]{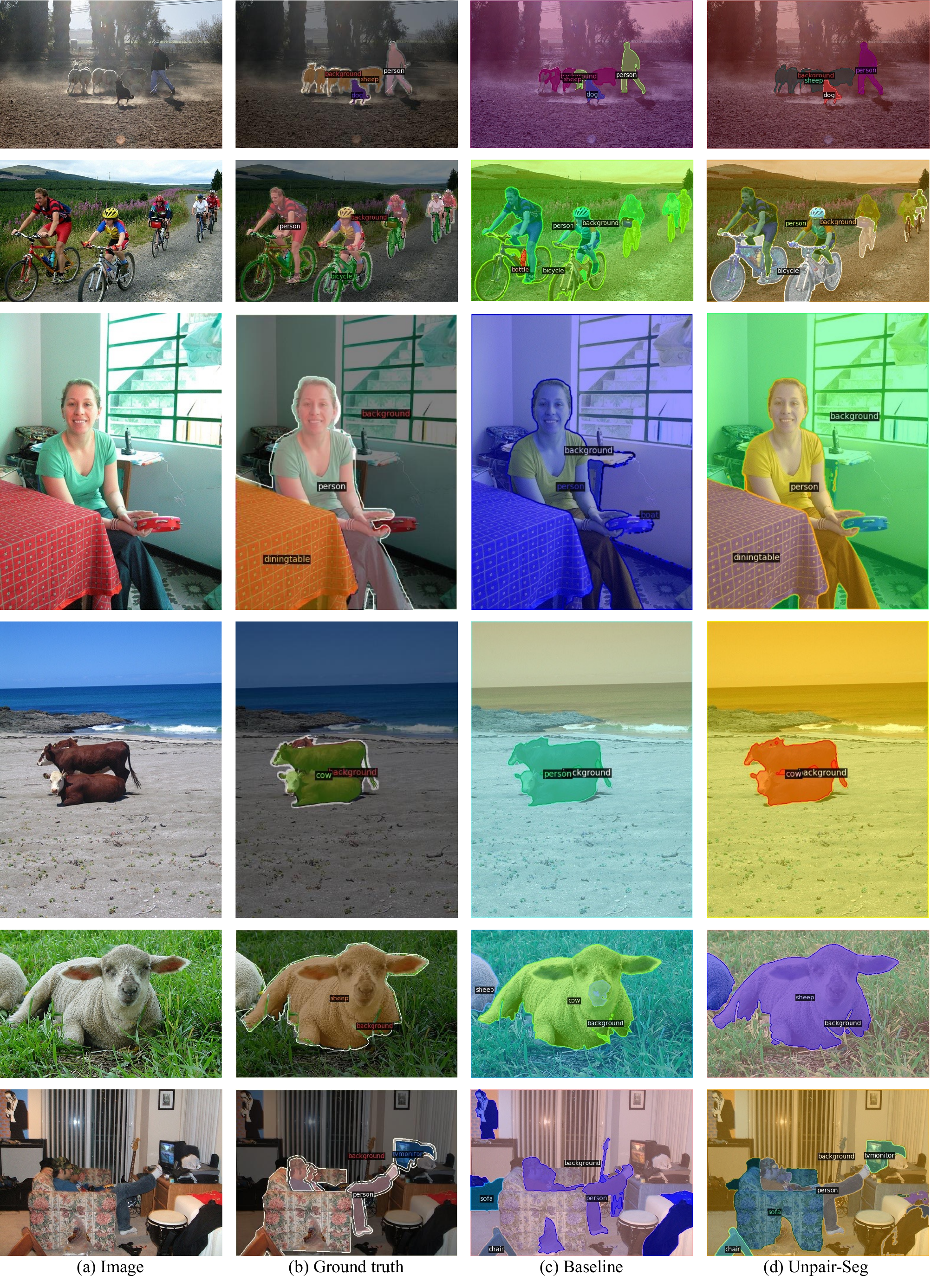}
    \vspace{-0.2cm}
    \caption{Visualisation of open-vocabulary segmentation between the baseline and Unpair-Seg.}
    \label{fig:p_sem_3}
    \vspace{-0.2cm}
\end{figure*}

\begin{figure*}[h]
    \centering
    \vspace{-0.2cm}
    \includegraphics[width=0.85\textwidth]{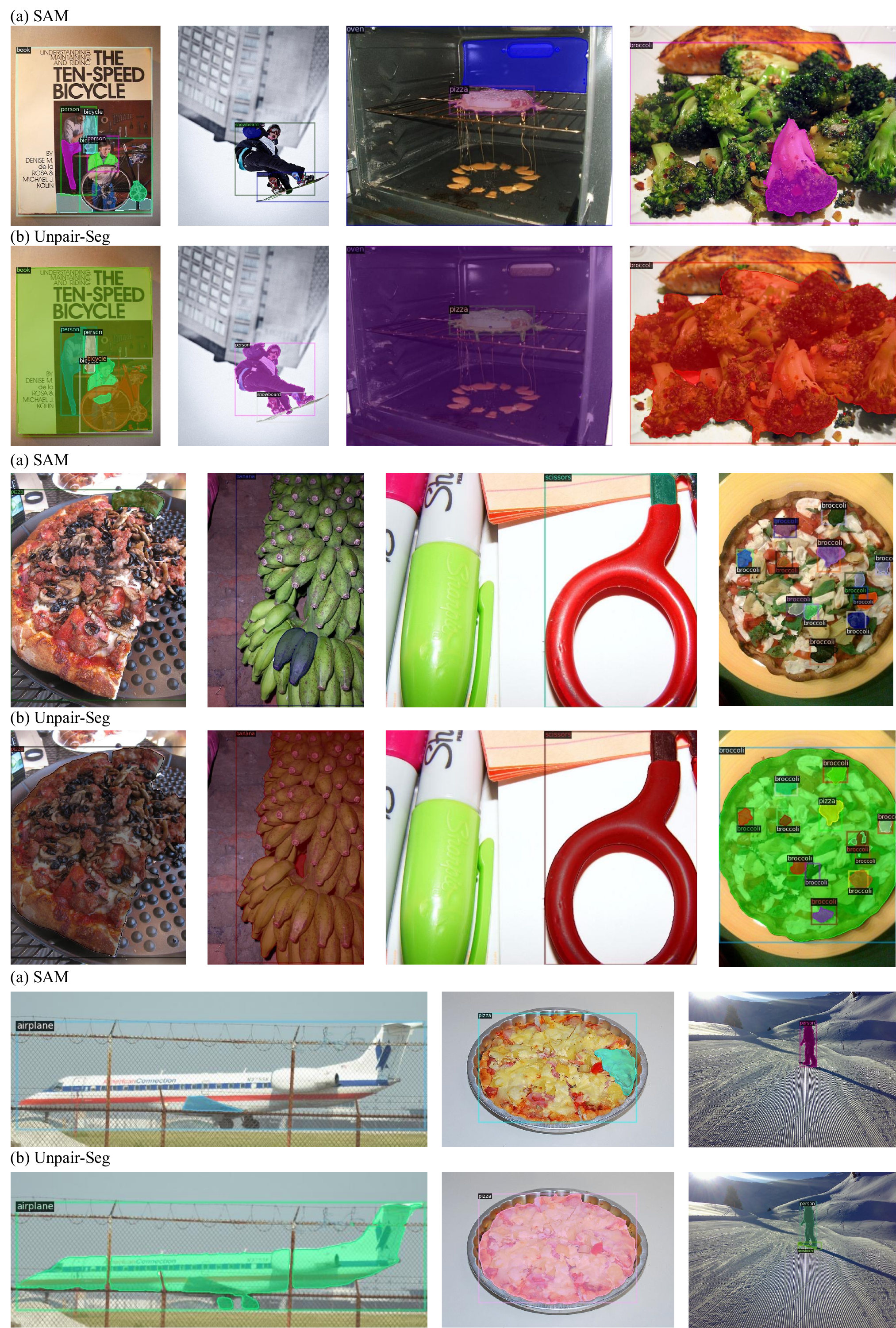}
    \vspace{-0.2cm}
    \caption{Visualisation of promptable segmentation between SAM-ViT/L and Unpair-Seg.}
    \label{fig:p_seg_1}
    \vspace{-0.2cm}
\end{figure*}
\begin{figure*}[h]
    \centering
    \vspace{-0.2cm}
    \includegraphics[width=0.85\textwidth]{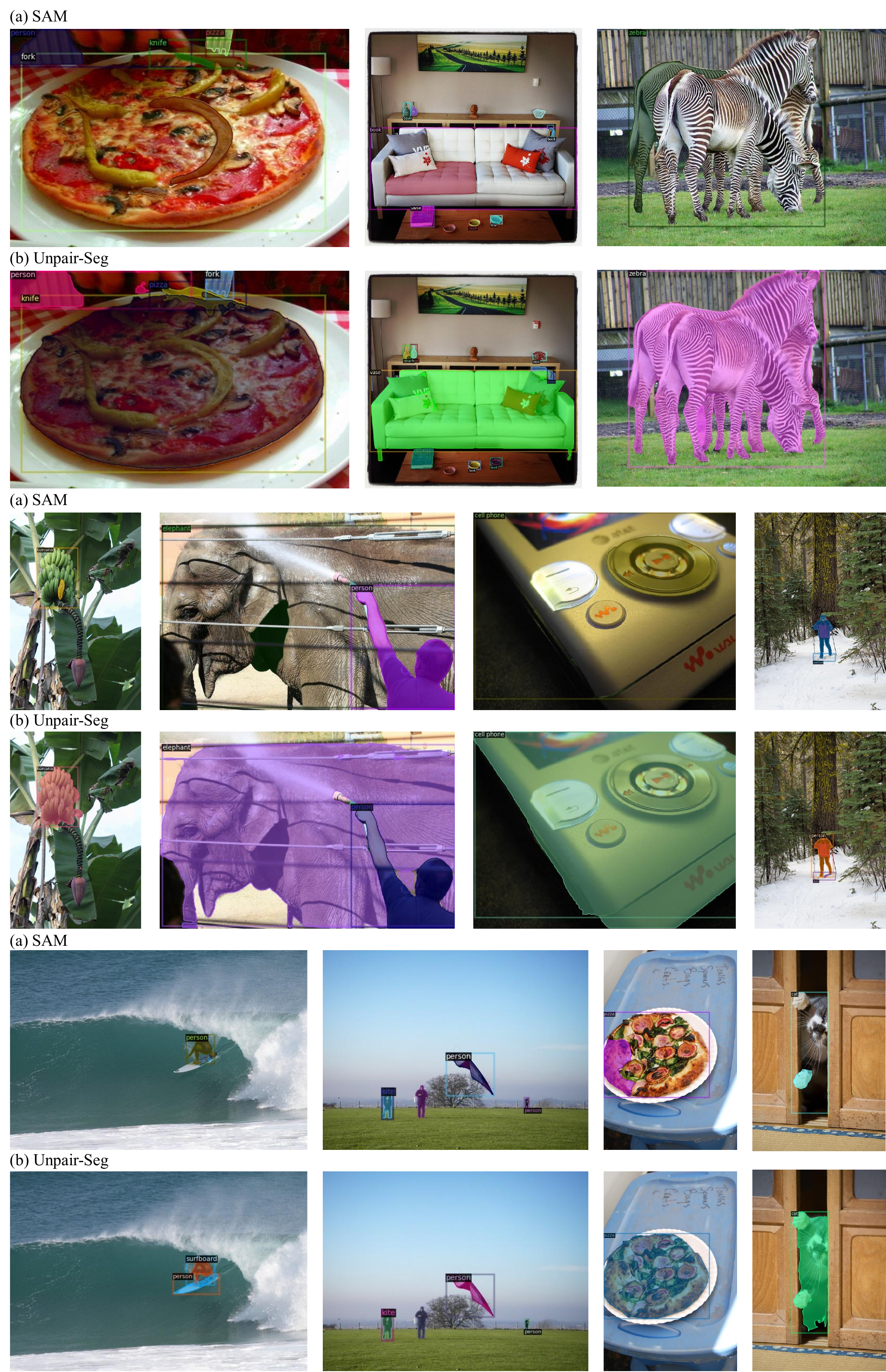}
    \vspace{-0.2cm}
    \caption{Visualisation of promptable segmentation between SAM-ViT/L and Unpair-Seg.}
    \label{fig:p_seg_2}
    \vspace{-0.2cm}
\end{figure*}
\begin{figure*}[h]
    \centering
    \vspace{-0.2cm}
    \includegraphics[width=0.85\textwidth]{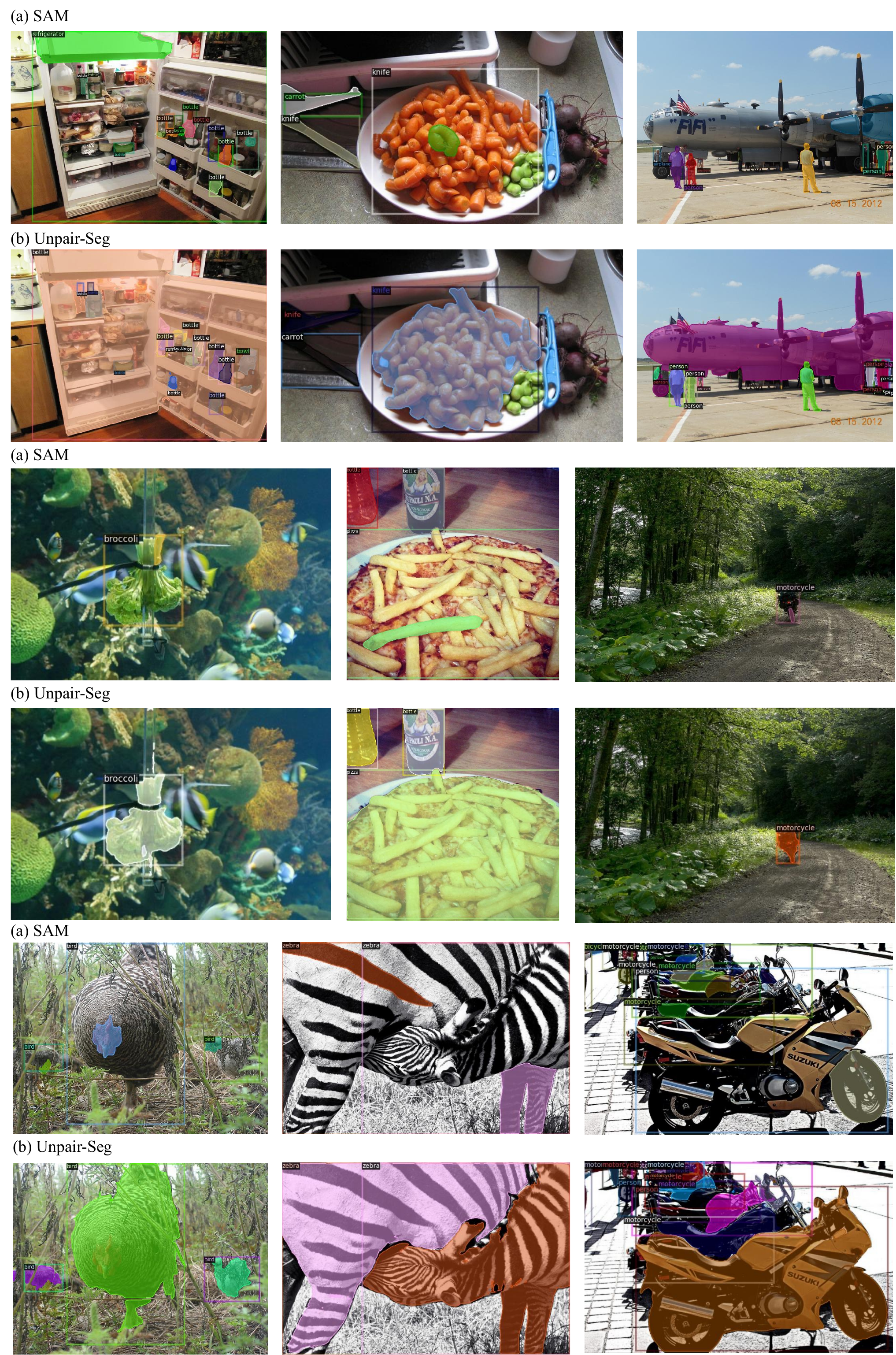}
    \vspace{-0.2cm}
    \caption{Visualisation of promptable segmentation between SAM-ViT/L and Unpair-Seg.}
    \label{fig:p_seg_3}
    \vspace{-0.2cm}
\end{figure*}
\begin{figure*}[h]
    \centering
    \vspace{-0.2cm}
    \includegraphics[width=0.85\textwidth]{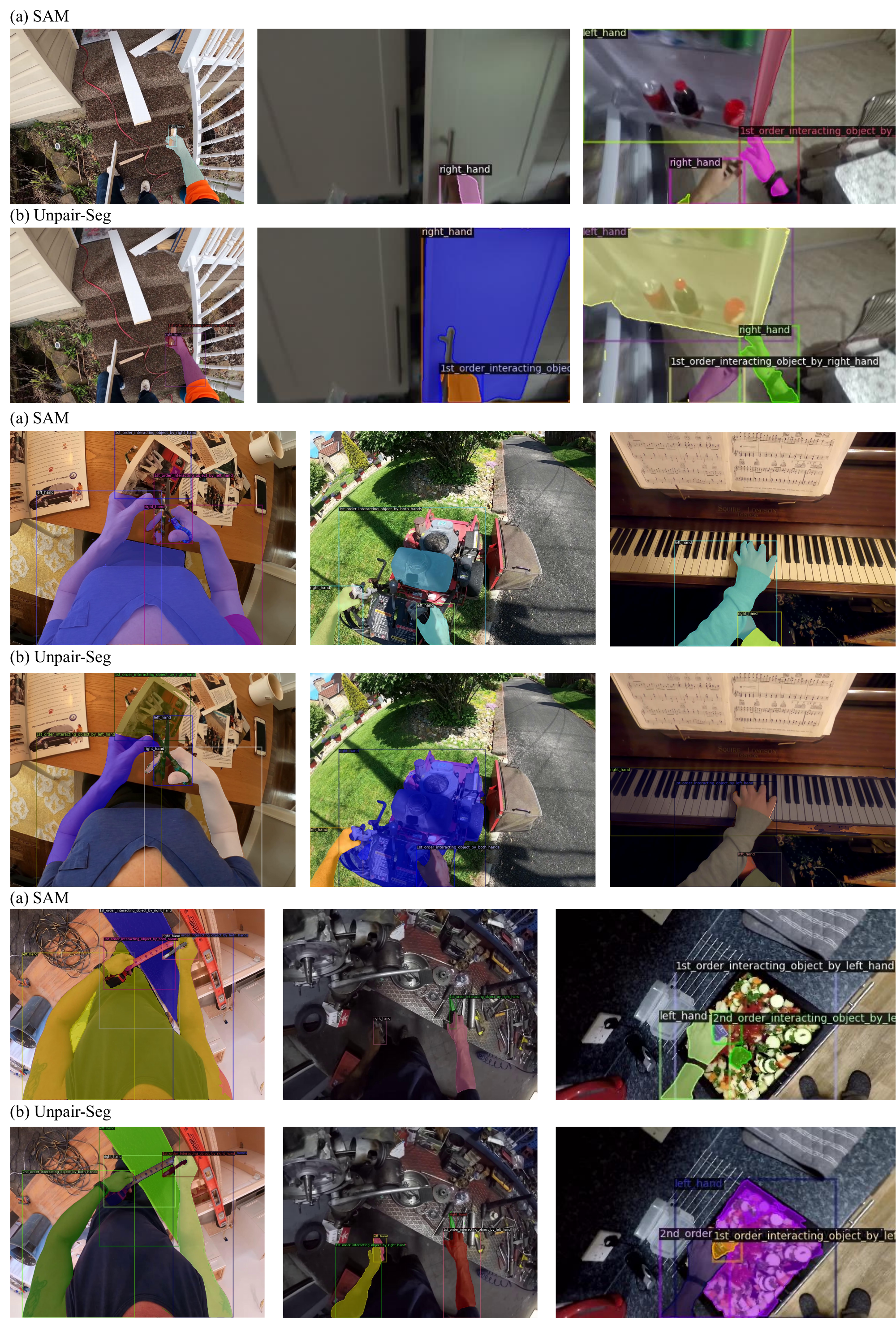}
    \vspace{-0.2cm}
    \caption{Visualisation of promptable segmentation between SAM-ViT/L and Unpair-Seg.}
    \label{fig:p_seg_4}
    \vspace{-0.2cm}
\end{figure*}
\begin{figure*}[h]
    \centering
    \vspace{-0.2cm}
    \includegraphics[width=0.85\textwidth]{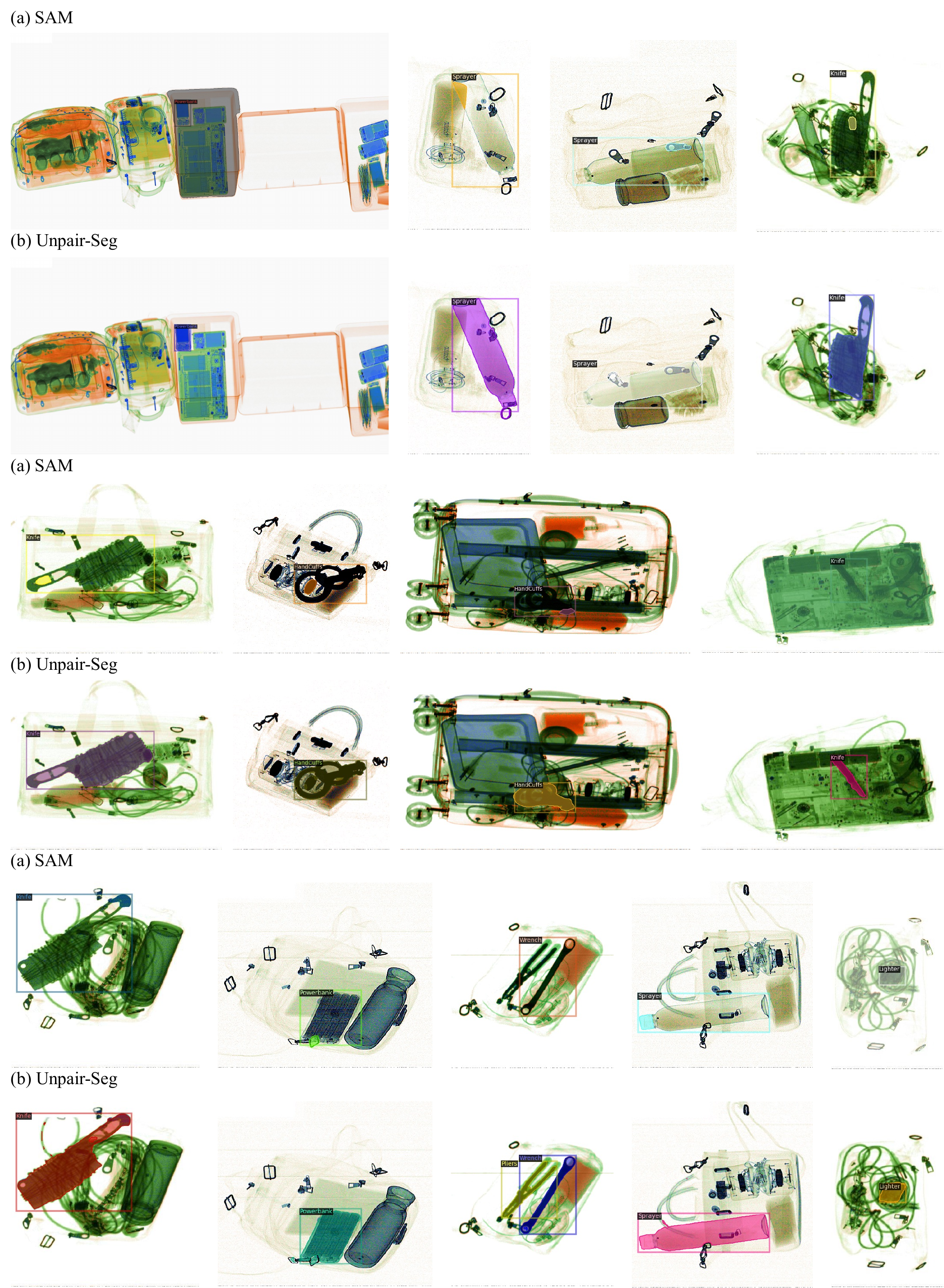}
    \vspace{-0.2cm}
    \caption{Visualisation of promptable segmentation between SAM-ViT/L and Unpair-Seg.}
    \label{fig:p_seg_5}
    \vspace{-0.2cm}
\end{figure*}

%%%%%%%%%%%%%%%%%%%%%%%%%%%%%%%%%%%%%%%%%%%%%%%%%%%%%%%%%%%%%%%%%%%%%%%%%%%%%%%
%%%%%%%%%%%%%%%%%%%%%%%%%%%%%%%%%%%%%%%%%%%%%%%%%%%%%%%%%%%%%%%%%%%%%%%%%%%%%%%

% \clearpage
% \input{sections/09_checklist}
\end{document}